\documentclass{article}

\usepackage{microtype}
\usepackage{graphicx}
\usepackage{booktabs} 

\usepackage{physics}
\usepackage{amsmath}
\usepackage[cjk]{kotex}
\usepackage{multirow}
\usepackage{booktabs}
\usepackage{tabularx}
\usepackage{colortbl}
\usepackage{graphicx}
\usepackage{bbding}
\usepackage{pifont}
\usepackage{wasysym}
\usepackage{amssymb}
\usepackage{pdfrender}
\usepackage{wrapfig}
\usepackage{subcaption}
\usepackage{caption}
\usepackage{scalerel}
\usepackage{bbm}
\usepackage{xcolor}
\usepackage{url}
\usepackage{arydshln}

\newcommand\clearrow{\global\let\rowmac\relax}
\newcommand{\specialcell}[2][c]{
  \begin{tabular}[#1]{@{}c@{}}#2\end{tabular}}

\clearrow
\newcommand{\xmark}{\ding{55}}

\newcommand{\mshort}{iMHSA}

\usepackage{hyperref}

\usepackage[accepted]{icml2024}

\usepackage{amsmath}
\usepackage{amssymb}
\usepackage{mathtools}
\usepackage{amsthm}

\usepackage[capitalize,noabbrev]{cleveref}

\theoremstyle{plain}

\theoremstyle{definition}

\theoremstyle{remark}

\usepackage[textsize=tiny]{todonotes}
\usepackage{inconsolata}

\icmltitlerunning{Interactive Multi-Head Self-Attention with Linear Complexity}

\begin{document}

\twocolumn[
\icmltitle{Interactive Multi-Head Self-Attention with Linear Complexity}

\icmlsetsymbol{equal}{*}

\begin{icmlauthorlist}
\icmlauthor{Hankyul Kang}{ajou}
\icmlauthor{Ming-Hsuan Yang}{google,uc_merced}
\icmlauthor{Jongbin Ryu}{ajou}
\end{icmlauthorlist}

\icmlaffiliation{ajou}{Ajou University}
\icmlaffiliation{google}{Google}
\icmlaffiliation{uc_merced}{University of California at Merced}

\icmlcorrespondingauthor{Jongbin Ryu}{jongbin.ryu@gmail.com}

\vskip 0.3in
]

\makeatletter\def\Hy@Warning#1{}\makeatother
\printAffiliationsAndNotice{}  
\begin{abstract}
We propose an efficient interactive method for multi-head self-attention via decomposition. 
For existing methods using multi-head self-attention, the attention operation of each head is computed independently.
However, we show that the interactions between cross-heads of the attention matrix enhance the information flow of the attention operation.
Considering that the attention matrix of each head can be seen as a feature of networks, it is beneficial to establish connectivity between them to capture interactions better.
However, a straightforward approach to capture the interactions between the cross-heads is computationally prohibitive as the complexity grows substantially with the high dimension of an attention matrix.
In this work, we propose an effective method to decompose the attention operation into query- 
and key-less components.
This will result in a more manageable size for the attention matrix, specifically for the cross-head interactions.
Expensive experimental results show that the proposed cross-head interaction approach performs favorably against existing efficient attention methods and state-of-the-art backbone models.
\end{abstract}

\section{Introduction}

\begin{figure}[t]
    \centering
    \hfill
    \begin{subfigure}[b]{0.18\paperwidth}
        \includegraphics[width=\linewidth]{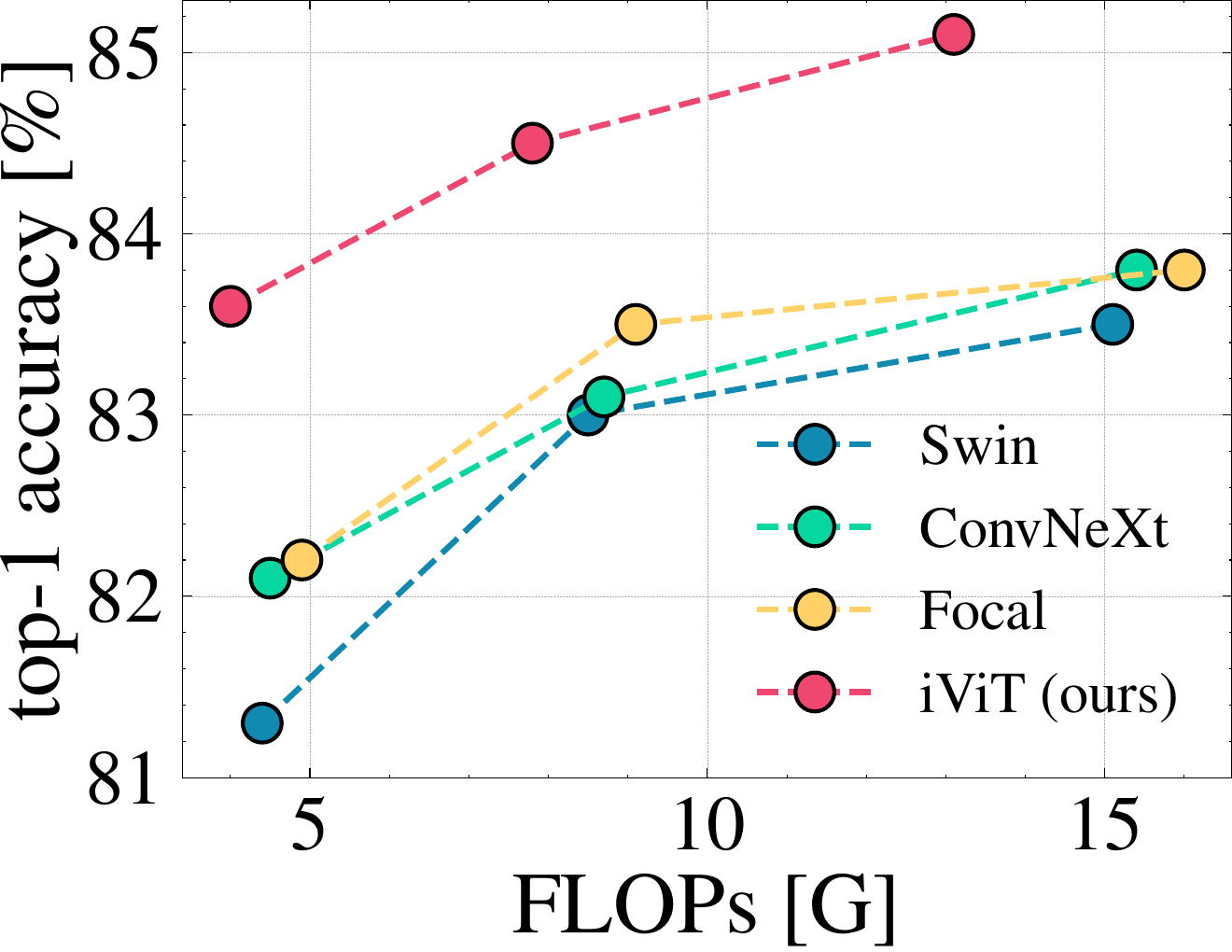}
        \subcaption{Classification}
    \end{subfigure}
    \hfill
    \begin{subfigure}[b]{0.18\paperwidth}
       \includegraphics[width=\linewidth]{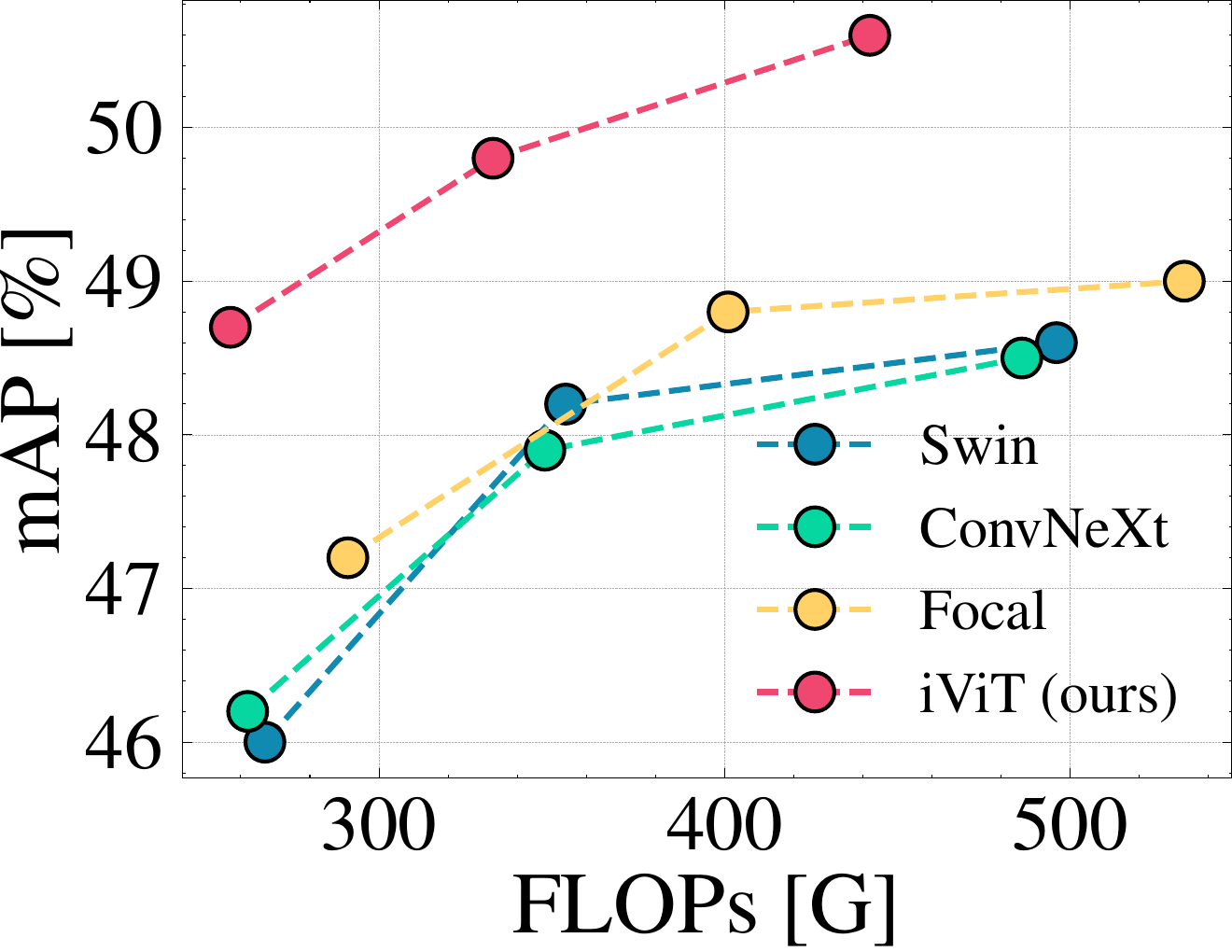}
       \subcaption{Detection}
    \end{subfigure}
    \hfill

    \hfill
    \begin{subfigure}[b]{0.18\paperwidth}
       \includegraphics[width=\linewidth]{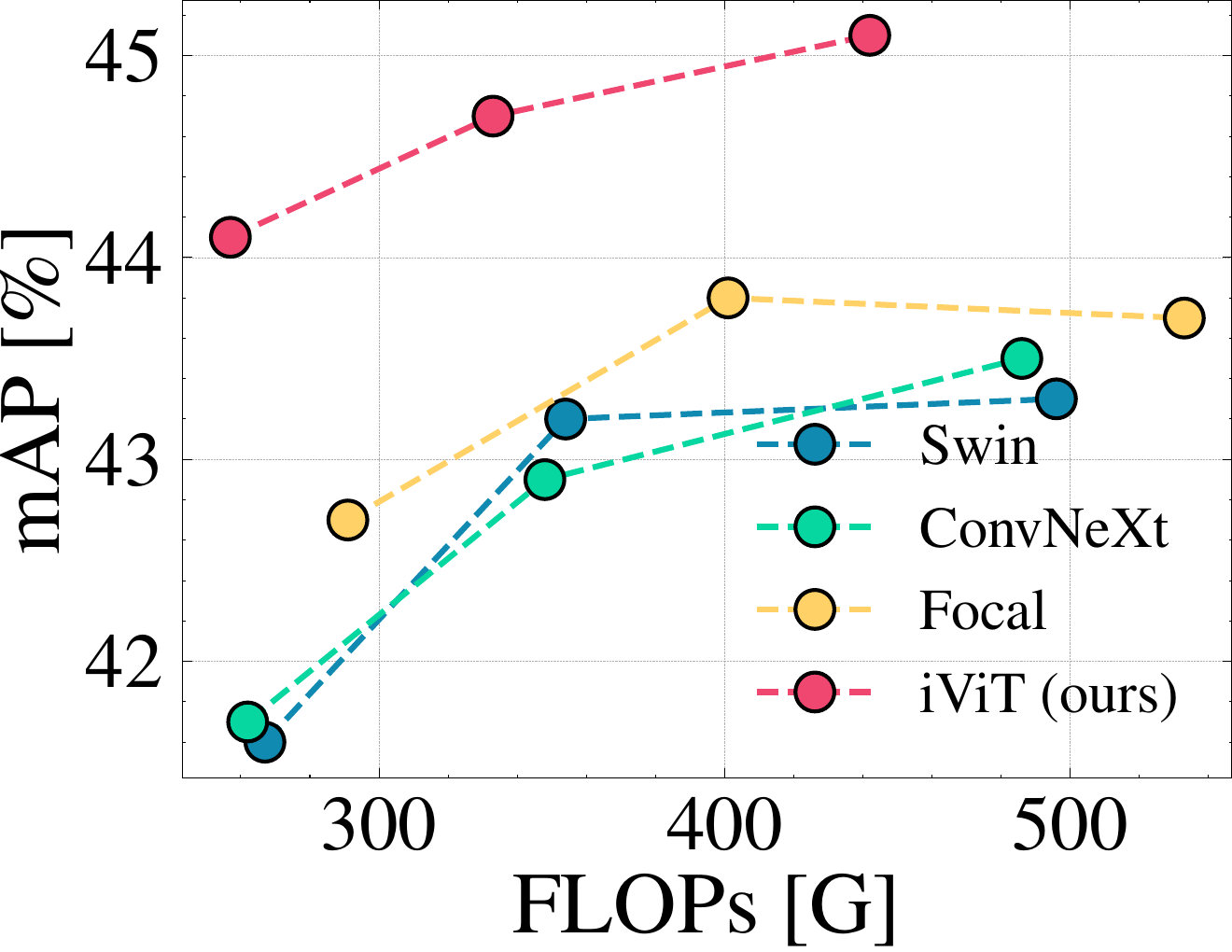}
       \subcaption{Instance segmentation}
    \end{subfigure}
    \hfill
    \begin{subfigure}[b]{0.18\paperwidth}
        \includegraphics[width=\linewidth]{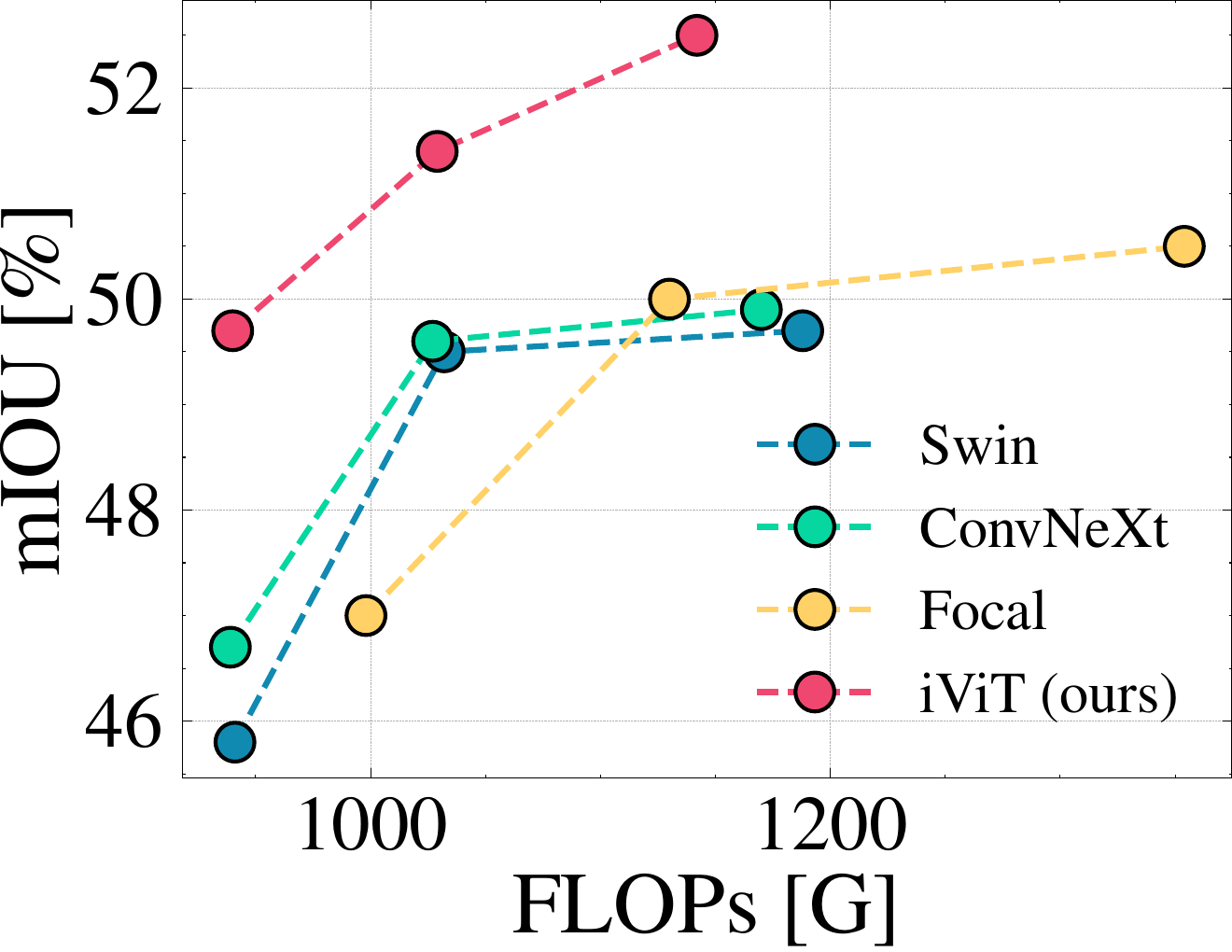}
        \subcaption{Semantic segmentation}
    \end{subfigure}
    \hfill
    \caption{
        Experimental evaluations of our and SOTA networks. 
        We compare the trade-off between computational complexity (FLOPs) and performance (top-1 accuracy, mAP, and more) on four tasks. 
    }
    \label{fig:plot_flops_performance}
\end{figure}

Establishing strong interactions between features of the networks is crucial in deep models~\cite{wang2017residual,hu2018squeeze,woo2018cbam}. 
Increased interactions on features facilitate the collection of diverse perspectives on the data, enhancing the ability to solve complex problems. 
As such, the key factor in constructing the effective deep neural network architecture is typically learning rich feature correlations by employing dense interactions~\cite{vaswani2017attention,dosovitskiy2020image}.
Nevertheless, it is infeasible to capture such interactions
in multi-head attention operations
due to the dimension of the attention matrix.
The dimension of the attention matrices renders it infeasible to employ the interaction layers. 
Moreover, when there is no interaction between heads, it is shown that performance reaches a plateau quickly, even with few heads.

In this work, we propose to reduce the dimension of the attention matrix to allow for cross-head interaction within the linear computational complexity.
We separate the self-attention operation into key- and query-less components by our decomposition approach.
Each component requires only the linear computational complexity so that we reduce the entire self-attention operation significantly. 
In addition, we employ a reverse-order attention computation to decrease the dimension of the attention matrix by altering the ordinary order of self-attention operation.
Hence, the proposed self-attention decomposition method accelerates the cross-head interaction computation and significantly diminishes memory usage.
These capabilities are crucial for the following reasons: 1) They enhance self-attention performance by facilitating interaction between cross-heads. 
2) They allow self-attention operations to be applied to large token sizes without exhausting memory usage. 
3) They make it possible to use self-attention in the early stages of attention blocks.
These benefits of our interactive multi-head self-attention lead us to go beyond simply improving the performance of self-attention and establishing a new strategy for designing vision transformer architecture.
We introduce a new hierarchical vision transformer architecture that achieves state-of-the-art performance for various visual recognition tasks, as shown in \Cref{fig:plot_flops_performance}.

\section{Related Work}
\paragraph{Sparsity-based attention.} Numerous methods have recently been developed to extensively analyze the sparsity of the attention matrix to avoid unnecessary computation. 
In \cite{child2019generating}, the attention region 
is restricted by a pre-defined local window to reduce computation complexity. 
Since this simple window-based splitting strategy ignores the similarity between tokens, 
more sophisticated splitting strategies \cite{kitaev2020reformer, chen2021scatterbrain}
are developed based on a locality-sensitive hashing (LSH) algorithm~\cite{indyk1998approximate}. 
In addition, a method that 
alternates between local and grid attention operations to enhance global information have been proposed \cite{tu2022maxvit}. 

\paragraph{Kernel-based attention.} Recent methods explore efficient attention modeling using non-linear kernels without directly computing the full attention matrix. 
For example, several schemes compute the output tokens in the reverse sequence of value, key, and query after applying a non-linear activation function to the query and key
\cite{katharopoulos2020transformers, shen2021efficient,choromanski2020rethinking}.
This reverse-order approach leads to linear complexity due to removing a quadratic attention matrix while approximating the non-linear activation of the softmax. 
To improve the approximation quality,
Han et al. 
propose focused linear attention to enhance dissimilarity across tokens via computing element-wise power of query and key \cite{han2023flatten}.
In addition, Cai et al. introduce a multi-branch approach
\cite{cai2023efficientvit}, in which each branch conducts linear attention with different dimensions of key and value.

\paragraph{Low-rank-based attention.} 
Numerous low-rank approximation schemes have been applied to reduce the dimensions of the attention matrix for multiple tasks.
In \cite{wang2020linformer,wang2021pyramid,wang2022pvt},
key and value are projected to a lower dimensional space via the pooling method.
On the other hand, tokens are clustered into a predetermined number of landmarks
using the LSH algorithm \cite{vyas2020fast}.
Furthermore, Zhu et al. project key and value by
multiplying a transposed matrix, each of which yields an asymmetric low-rank matrix. 
However, this asymmetric low-rank attention matrix inevitably loses information due to an incomplete attention matrix. 
To minimize the information loss, Xiong et al. approximate the attention matrix \cite{xiong2021nystromformer} using singular value decomposition (SVD). 
They estimate the Moore-Penrose inverse matrix with iterative matrix multiplications to speed up the SVD computation. 
Recently, Zheng et al. select landmarks through self-normalized importance sampling (SNIS) to restrict the attention operation around only the neighborhood tokens of a landmark \cite{zheng2023efficient}.

\paragraph{Refined attention.}
A few approaches aim to refine the attention matrix to capture more correlation between tokens.
To enhance the correlation of local tokens, 
several methods combine the attention matrix with the positional encoding \cite{liu2021swin,dong2022cswin}.
Qin et al.  break down the cosine positional encoding into the query and key to preserve a linear attention form
\cite{qin2022cosformer}, and 
Huang et al. apply distinct orthogonal transformations on individual local groups to maintain locality during grid-attention operations \cite{huang2022orthogonal}. 
On the other hand, Zhou et al. 
integrate a normalization or linear projection layer with the attention matrix to mitigate the duplicated attention computations \cite{zhou2021refiner, zhou2021deepvit}.

\section{Method}
In this section, we present our efficient interactive Multi-Head Self-Attention (iMHSA) method as illustrated in 
\Cref{fig:concept_overview,fig:concept_scheme}. 
We first define the cross-head interaction for the multi-head self-attention (\Cref{sec:method1}). 
After introducing our attention decomposition, we apply the cross-head interaction to the decomposed self-attention matrix, yielding linear complexity (\Cref{sec:method2}). 
Finally, we instantiate our architectural design by exploiting the proposed efficient interactive attention (\Cref{sec:method3}). 

\subsection{Preliminaries}
\label{sec:method1}

\subsubsection{Multi-head self-attention}
We define the input tokens of $\vb{z} \in \mathbb{R}^{N \times c}$, query $\mathcal{Q} \in \mathbb{R}^{N \times c}$, key $\mathcal{K} \in \mathbb{R}^{N \times c}$, and value $\mathcal{V} \in \mathbb{R}^{N \times c}$ for the self-attention operation as
\begin{equation}
    \vb{z} = \{\vb*{z}_{i}|\vb*{z}_{i} \in \mathbb{R}^{c}\text{, and $1 \leq i \leq N$}\},
    \label{eq:input_tokens}
\end{equation}
\begin{equation}
    \mathcal{Q}, \mathcal{K}, \mathcal{V} = (W_{Q}, W_{K}, W_{V})\vb{z},
    \label{eq:dependent_var}
\end{equation}
where $W_{\mathcal{Q}}, W_{\mathcal{K}}, W_{\mathcal{V}} \in \mathbb{R}^{c \times c}$ are linear projection matrices.
Following the standard notation for the MHSA, we divide the channel dimension $c$ of the query, key, and value into the multi-head dimension $d$. We then calculate the output tokens per head $\vb*{z_i} \in \mathbb{R}^{N \times d}$ by multiplying the softmax-attention matrix $\mathcal{A} \in \mathbb{R}^{N \times N}$ with the value $\mathcal{V}$ as
\begin{equation}
        \text{$\vb*{t_i} = \mathcal{A}\mathcal{V}$, where $\mathcal{A} = \mathrm{softmax}(\mathcal{Q} \mathcal{K}^{\top}/d^{\frac{1}{2}})$}.
    \label{eq:softmax_attention}
\end{equation}
We concatenate all output tokens from multiple heads to yield the final output tokens $\vb{z} \in \mathbb{R}^{N \times c}$ as
\begin{equation}
    \vb{z} = \underset{\mathrm{ch}}{\mathrm{concat}}(\{\vb*{z}_{i}|\vb*{z}_{i} \in \mathbb{R}^{N \times d}\text{; and $1 \leq i \leq H$}\}),
    \label{eq:multi_head_softmax_attention}
\end{equation}
where $H$ denotes the number of heads.

\begin{figure}
    \centering
    
    \begin{subfigure}[b]{0.18\paperwidth}
        \centering        
        \includegraphics[width=\linewidth]{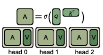}
        \caption{MHSA}
        \label{fig:concetp_overview_attention}
    \end{subfigure}
    \begin{subfigure}[b]{0.18\paperwidth}
        \centering        
        \includegraphics[width=\linewidth]{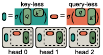}
        \caption{MHSA w/ Decomposition}
        \label{fig:concetp_overview_decomposed_attention}
    \end{subfigure}
    
    \begin{subfigure}[b]{0.18\paperwidth}
        \centering        
        \includegraphics[width=\linewidth]{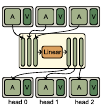}
        \caption{MHSA w/ Interaction}
        \label{fig:concetp_overview_attention_with_interaction}
    \end{subfigure}
    \begin{subfigure}[b]{0.18\paperwidth}
        \centering        
        \includegraphics[width=\linewidth]{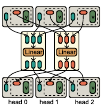}
        \caption{iMHSA (ours)}
        \label{fig:concetp_overview_decomposed_attention_with_interaction}
    \end{subfigure}
    
    \caption{
    Schematic illustration of the (a) baseline MHSA, (b) MHSA with only Decomposition, (c) MHSA with only Interaction, and (d) our iMHSA with both Decomposition and Interaction.
    }
    \label{fig:concept_overview}
\end{figure}

\begin{figure}[t]
    \centering
    \begin{subfigure}[b]{0.36\paperwidth}
        \centering        
        \includegraphics[width=\linewidth]{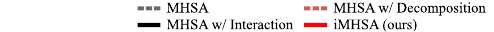}
    \end{subfigure}
    
    \begin{subfigure}[b]{0.36\paperwidth}
        \centering        
        \includegraphics[width=0.49\linewidth]{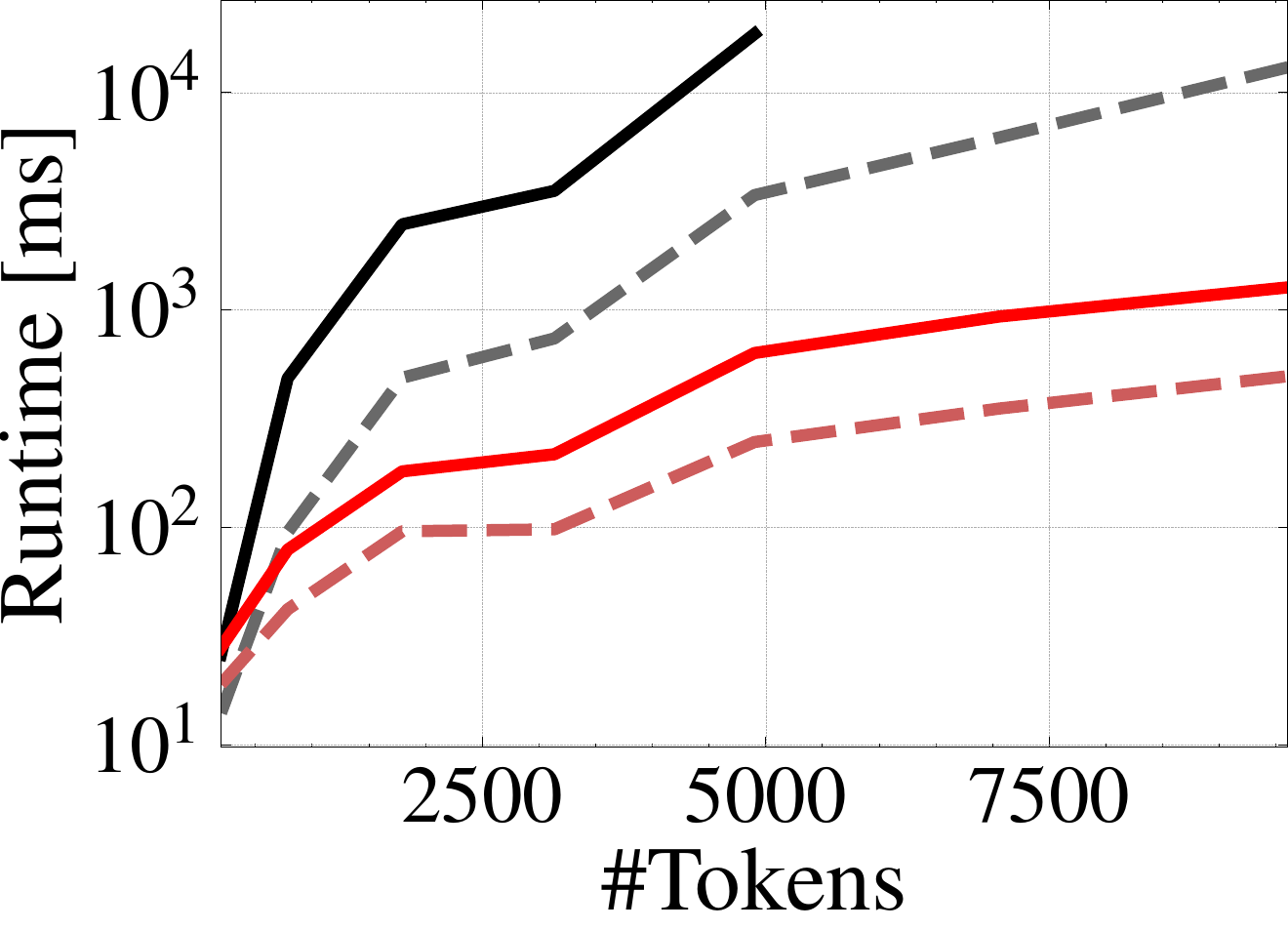}
        \includegraphics[width=0.49\linewidth]{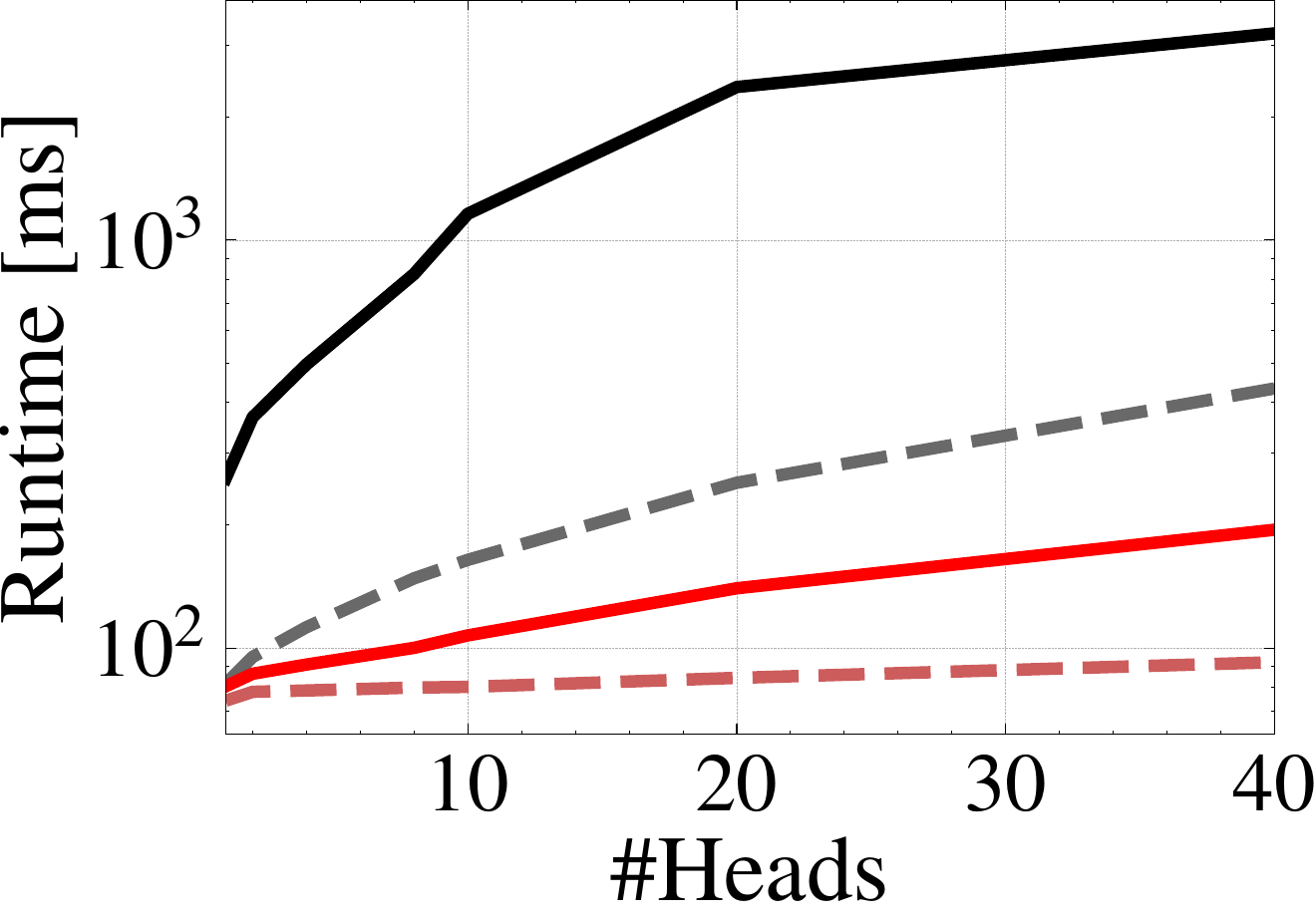}
        \caption{Runtime}
        \label{fig:plot_runtime}
    \end{subfigure}
    \begin{subfigure}[b]{0.36\paperwidth}
        \centering        
        \includegraphics[width=0.49\linewidth]{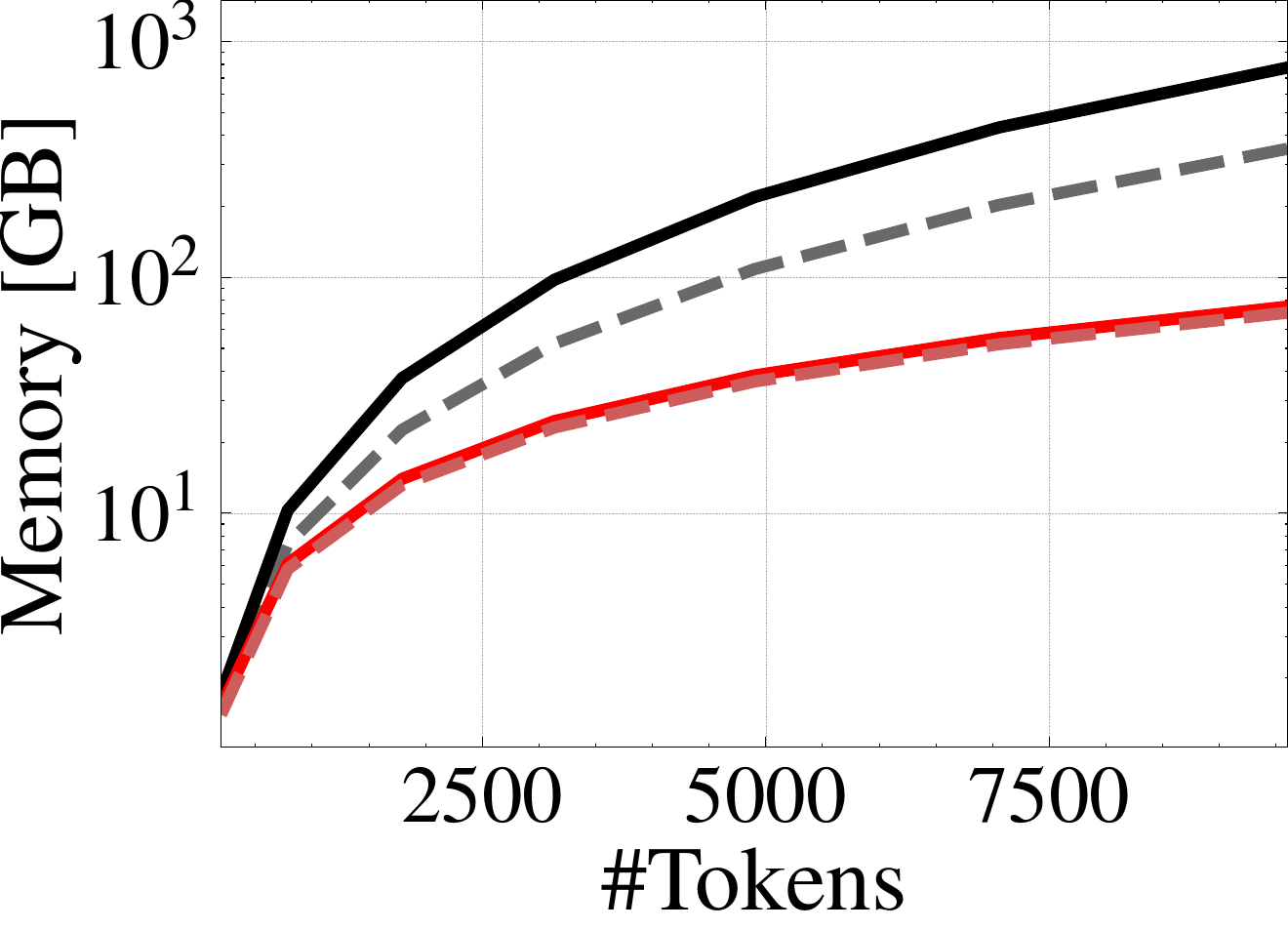}
        \includegraphics[width=0.49\linewidth]{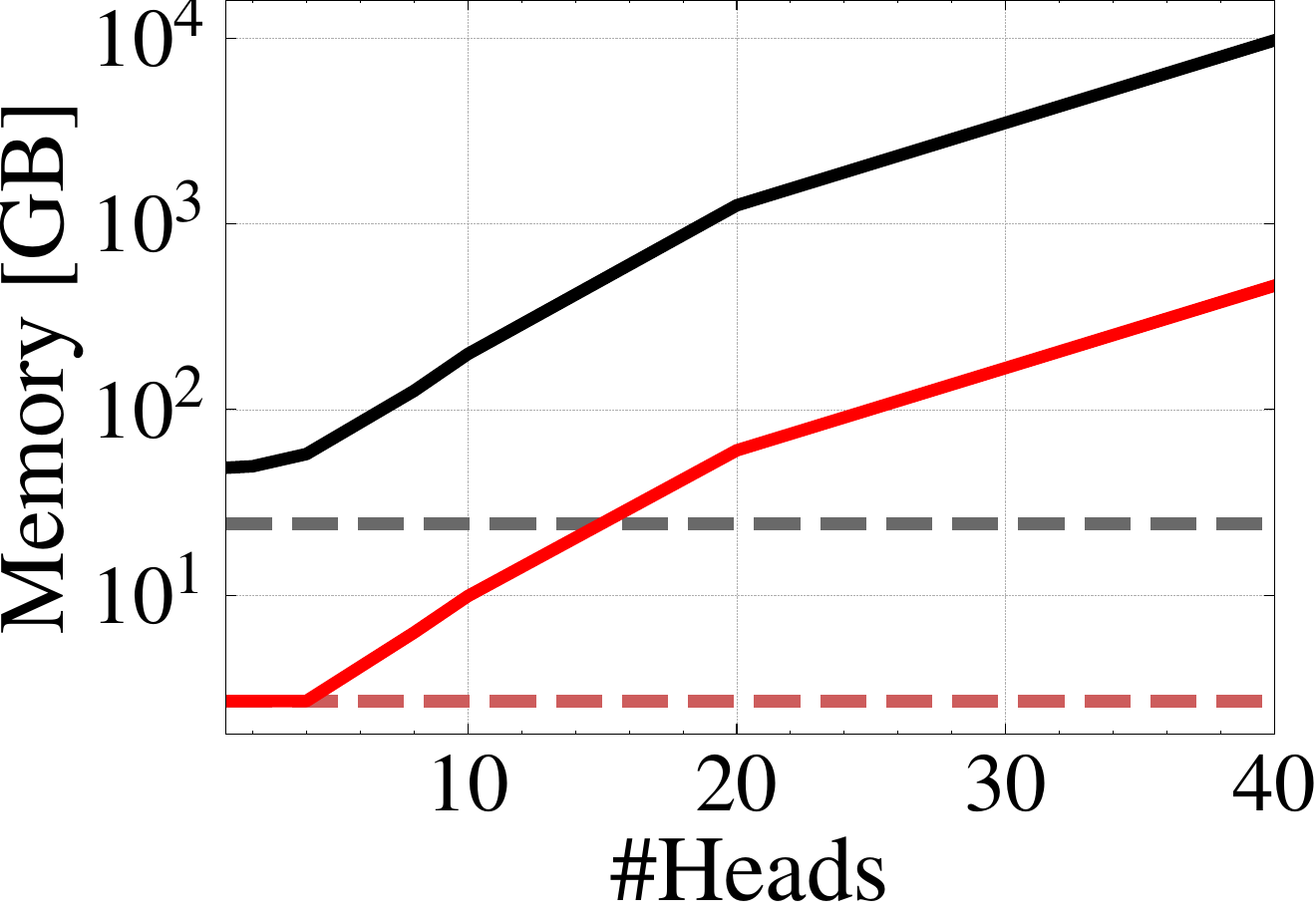} 
        \caption{Memory usage}
        \label{fig:plot_memory}
    \end{subfigure}
    \caption{
    Experimental comparisons of the runtime and memory usage. We measure them in a single attention block using ViT-S. The `MHSA with the cross-head Interaction' enormously increases the runtime and memory usage, but our Decomposition method requires minimal resources even with the Interaction. The `iMHSA' denotes our approach with both the Interaction and Decomposition methods. 
    There is a very small difference in (b) between the `MHSA /w Decomposition' and `iMHSA'. 
    }
    \label{fig:plot_token_runtime_and_memory}
\end{figure}

\subsubsection{Cross-Head Interaction}
We propose a cross-head interaction method to improve the information flow between attention matrices of multiple heads. 
Since the output tokens are concatenated immediately after attention operation, as \Cref{eq:softmax_attention,eq:multi_head_softmax_attention}, there is no information exchange between attention matrices of different heads.   
Thus, one can suggest the cross-head interaction layers to enhance the information flow of attention matrices as
\begin{equation}
        \text{$\mathcal{A} = W_{2}\mathrm{softmax}(W_{1}\mathcal{S})$, where $\mathcal{S} = \mathcal{Q} \mathcal{K}^{\top}/d^{\frac{1}{2}}$,}
    \label{eq:cross_head_operation}
\end{equation}
where $W_{1}, W_{2} \in \mathbb{R}^{h \times h}$ are the fully-connected operation across head dimensions.

\paragraph{Computational Complexity.} 
While the cross-head interaction facilitates the information exchange, we must address an important issue requiring high computational complexity.
Since each of $H$ attention matrices has $\mathbb{R}^{N^2}$ dimension, the cross-head interaction layers induce extra computation of $O(N^2h^2)$, which significantly decreases the computational efficiency of a network as shown in
\cref{fig:plot_runtime}.

\paragraph{Memory Usage.} We also need to tackle the high memory usage due to the MHSA operation.
The memory issue is an inherent problem of the self-attention operation, but using the cross-head interaction layer exacerbates it. 
This is because the additional memory should be consumed by storing input and output attention maps for the cross-head interaction operation. 
As shown in \cref{fig:plot_memory}, the memory usage of a network is significantly increased by the cross-head interaction operation when the image size is enlarged.
Hence, considering these two issues, we propose the self-attention decomposition method to reduce the dimensional of attention matrices. 


  




\begin{figure}[t]
    \centering
    \includegraphics[width=\linewidth]{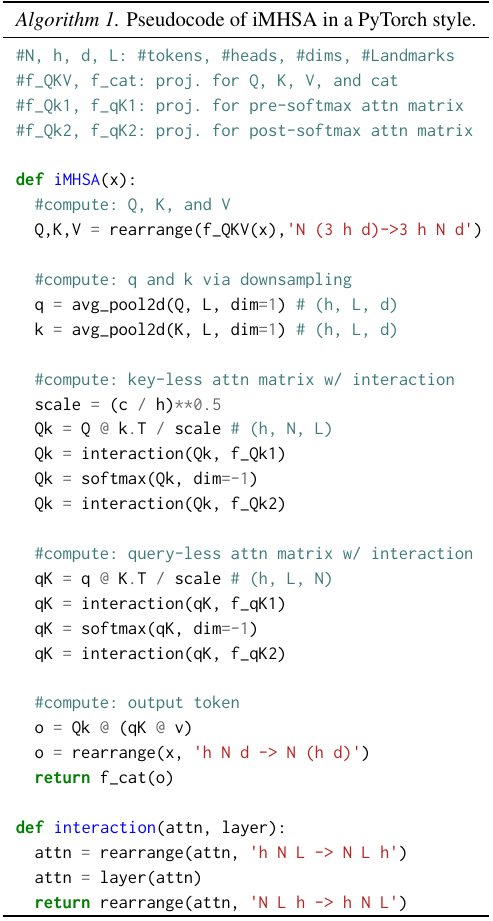}
    \label{fig:algorithm}
\end{figure}

\subsection{Interactive MHSA with Linear Complexity}
\label{sec:method2}

\subsubsection{Self-Attention Decomposition}
To cope with the high computational complexity and memory usage, the most straightforward approach is to downsize the dimension of the key and query.
However, simply reducing their dimension will result in the loss of discriminant information.
Not surprisingly, as the dimension increases, the network can learn more detailed information to achieve better performance.
Thus, we propose reducing the attention matrix size while maintaining the whole dimension of key and value. 
To accomplish this, we downsize the key and query independently to decompose two attention matrices, as shown in~\Cref{fig:concept_overview}. 
As such, despite key and value having less detailed information in each decomposed attention matrix, their full dimension is preserved in the opposite attention matrix.
We compute these decomposed attention matrices as 
\begin{align}
    \begin{split}
        q_{h,w} &= \frac{1}{\Omega_{h}\Omega_{w}}\sum\limits_{i=-\Omega_h}^{\Omega_h}\sum\limits_{j=-\Omega_w}^{\Omega_w}{Q_{h+i,w+j}}, \\
        k_{h,w} &= \frac{1}{\Omega_{h}\Omega_{w}}\sum\limits_{i=-\Omega_h}^{\Omega_h}\sum\limits_{j=-\Omega_w}^{\Omega_w}{K_{h+i,w+j}}, \\
        \mathcal{A} 
        &= \mathcal{A_{Q}}\mathcal{A_{K}} \\
        &= \mathrm{softmax}(\mathcal{Q}_{i}k^\top)
        \mathrm{softmax}(q\mathcal{K}^{\top}),
    \end{split}
    \label{eq:attn_weight_decomposition}
\end{align}
where $\Omega_{h},\Omega_{w}$ denote the kernel length of the downsampling method, $q, k \in \mathbb{R}^{L \times d}$ are landmarks of query and key with less information, and $L$ indicates the number of landmarks.
We use the average pooling to get the spatially reduced landmarks so that the query- and key-less attention matrices have significantly low dimensions as $\mathcal{A_{Q}}, \mathcal{A_{K}} \in \mathbb{R}^{N \times L}$ where $L << N$.

\begin{figure}[t]
    \centering
    
    \hfill
    \begin{subfigure}[b]{0.0955\paperwidth}
        \centering        
        \includegraphics[width=\linewidth]{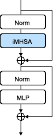}
        \caption{Block}
        \label{fig:concept_block}
    \end{subfigure}
    \hfill
    \begin{subfigure}[b]{0.2497\paperwidth}
        \centering        
        \includegraphics[width=\linewidth]{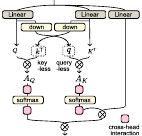}
        \caption{iMHSA}
        \label{fig:concept_imhsa}
    \end{subfigure}
    \hfill
    
    \begin{subfigure}[b]{0.36\paperwidth}
        \centering        
        \includegraphics[width=\linewidth]{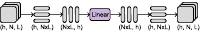}
        \caption{Cross-head interaction}
        \label{fig:concept_interaction}
    \end{subfigure}
    \caption{
    Architectural design of the (a) iMHSA-based attention block, (b) iMHSA, and (c) cross-head interaction. 
    }
    \label{fig:concept_scheme}
\end{figure}

\begin{table*}[t]
    \centering
    \small
    \renewcommand{\arraystretch}{1.2} 
    \setlength{\tabcolsep}{6pt}
    \caption{
    Main architectural design choice of our iViT in Tiny (T), Small (S), and Base (B) scales. 
    We set the MLP ratio to 4 and the number of landmarks to 49 at all scales. 
    We measure FLOPs with an image size of 224\textsuperscript{2}. 
    }
    \label{tab:architecture_configuration}
    \vskip 0.15in
    \begin{tabular}{@{}lrrrrrrr@{}}
        \toprule
        Scale & 
        \multicolumn{4}{c}{Block\textsuperscript{*} / \#Layers / \#Channels / \#Heads} &
        \multirow{2}{*}{\specialcell{\#P\\(M)}} & 
        \multirow{2}{*}{\specialcell{F\\(G)}} &
        \multirow{2}{*}{\specialcell{Speed\\(img/sec)}} \\
        \cmidrule{2-5}
        
        & Stage 1 & Stage 2 & Stage 3 & Stage 4 \\
        \midrule

        T & C / 3 / 64 & C / 3 / 128 & i / 9 / 320 / 10 & i / 3 / 512 / 16 & 
        27.5 & 4.0 & 1695 \\
        S & C / 3 / 64 & C / 12 / 128 & i / 18 / 320 / 10 & i / 3 / 512 / 16 & 
        41.1 & 7.8 & 930 \\
        B & C / 3 / 96 & C / 12 / 192 & i / 18 / 384 / 12 & i / 3 / 576 / 18 & 
        58.3 & 13.1 & 686 \\
        \bottomrule
        
        \multicolumn{8}{r}{
        Block\textsuperscript{*} type is either of \textbf{C}onvolution or \textbf{i}MHSA.
        } \\
        
    \end{tabular}
    \vskip -0.1in
\end{table*}

\subsubsection{Interactive MHSA with Linear Complexity}
We introduce the cross-head interaction method, as shown in \Cref{fig:concept_scheme}, performing cross-head interactions on the decomposed-attention matrices.
We use cross-head interaction of \Cref{eq:cross_head_operation} to $\mathcal{A_Q}, \mathcal{A_K}$ in \Cref{eq:attn_weight_decomposition} as
\begin{align}
    \begin{split}
    \mathcal{A} &= \mathcal{A_{Q}}\mathcal{A_{K}}, \\
    \mathcal{A_{Q}} &= W_{2}^{\mathcal{Q}}\mathrm{softmax}(W_{1}^{\mathcal{Q}}\mathcal{Q} k^{\top}/d^{\frac{1}{2}}), \\
    \mathcal{A_{K}} &= W_{2}^{\mathcal{K}}\mathrm{softmax}(W_{1}^{\mathcal{K}}q \mathcal{K}^{\top}/d^{\frac{1}{2}}),
    \end{split}
    \label{eq:linearly_interactive}
\end{align}
where $W_{1}^{\mathcal{Q}}, W_{2}^{\mathcal{Q}}, W_{1}^{\mathcal{K}}, W_{2}^{\mathcal{K}} \in \mathbb{R}^{h \times h}$ refer to cross-head interaction layers. $h$ stands for the channel dimension.

To compute the output token, we reorder multiplications using the commutativity of matrix multiplication to avoid directly computing $\mathcal{A_Q}\mathcal{A_K}^\top$ as 
\begin{align}
    \begin{split}
    \vb*{o} 
    = (\mathcal{A_{Q}}\mathcal{A_{K}}^\top)\mathcal{V}
    = \mathcal{A_{Q}}(\mathcal{A_{K}}^\top\mathcal{V}).
    \end{split}
    \label{eq:linearly_interactive_sub}
\end{align}

\paragraph{Computational Complexity.} In our~\mshort, cross-head interaction only increases computation by $O(2NLh^2)$ due to reduced attention matrix. 
Furthermore, our decomposition method does not directly access the full attention matrix. Therefore, our~\mshort~has linear complexity of $O(2NLh(d+h))$, making it suitable for large token sizes.

\paragraph{Memory Usage.} 
Thanks to the commutative operation described in Eq.\ref{eq:linearly_interactive_sub}, we do not need to compute the full attention matrix $\mathcal{A} \in \mathbb{R}^{N^2}$. 
Instead, we only require the intermediate attention matrix $\mathcal{A_{K}}^\top\mathcal{V}$ with a manageable dimension of $\mathbb{R}^{N \times h}$.
This is because the channel dimension $h$ remains constant while the token size $N$ changes depending on the input image size in the decomposition method. 
Furthermore, in the earlier stages of a pyramidal network backbone, the memory usage grows linearly because $h << N$.

The merits of the proposed method in terms of computational complexity and memory usage can be found in \Cref{fig:plot_token_runtime_and_memory}.
The pseudocode in Algorithm 1 summarizes the main steps of the proposed method.

\subsection{Interactive Vision Transformer}
\label{sec:method3}
We present a new network backbone to demonstrate the advantages of our proposed iMHSA. 
Based on the recently proposed PoolFormer~\cite{yu2022metaformer}, we apply our iMHSA in Stages 3 and 4 as shown in~\cref{tab:architecture_configuration}. We refer to our network as an interactive Vision Transformer (iViT). Compared to other SOTA networks, our iViT works favorably with few resources for various vision tasks.

\section{Experimental Results}
\subsection{Evluation against Efficient Attention Methods}
We evaluate the proposed iMHSA algorithm and other efficient attention methods such as Softmax~\cite{vaswani2017attention}, Performer~\cite{choromanski2020rethinking}, Local attention~\cite{child2019generating}, Scatterbrain~\cite{chen2021scatterbrain}, Nystr\"omformer~\cite{xiong2021nystromformer}, LARA~\cite{zheng2022linear}, Combiner~\cite{ren2021combiner}, Long-Short~\cite{zhu2021long}, and EVA~\cite{zheng2023efficient}.

We perform extensive comparisons under various input image resolutions on the ImageNet-1k dataset.
For fair comparisons, we use the same training recipe of previous work~\cite{zheng2023efficient} with 300 epochs. 
%

\begin{table}[t]
    \centering
    \scriptsize
    \renewcommand{\arraystretch}{1.1} 
    \setlength{\tabcolsep}{5pt}
    \caption{Experimental results on standard token size (\#token=196). 
    We compare our ViT with other efficient attention methods on standard Vision Transformer networks with an image size of $224^2$. 
    }
    \label{tab:baseline_networks}
    \vskip 0.15in
    \begin{tabular}{@{}lrrrrrrr@{}}
        \toprule
        
        Methods
        & \multicolumn{3}{c}{ViT-T/16} &
        & \multicolumn{3}{c}{ViT-S/16} \\
        \cmidrule(lr){2-4}
        \cmidrule(lr){6-8}

        & 
        \specialcell{\#Param.\\(M)} 
        & \specialcell{FLOPs\\(G)} 
        & \specialcell{Top-1\\Acc.(\%)} & &
        \specialcell{\#Param.\\(M)} &
        \specialcell{FLOPs\\(G)} & 
        \specialcell{Top-1\\Acc.(\%)} \\
        \midrule

        Softmax & 5.7 & 1.3 & 73.0 & & 22.0 & 4.6 & 80.4 \\ \hdashline[1pt/1pt] 
        Local & 5.7 & 1.1 & 67.1 & & 22.0 & 4.3 & 74.1 \\ \hdashline[1pt/1pt] 
        Performer & 5.7 & 1.2 & 65.9 & & 22.0 & 4.4 & 74.3 \\ \hdashline[1pt/1pt] 
        LARA & 5.8 & 1.2 & 71.5 & & 22.2 & 4.5 & 79.5 \\ \hdashline[1pt/1pt] 
        EVA & 5.8 & 1.2 & 73.0 & & 22.2 & 4.4 & 80.7 \\ \hdashline[1pt/1pt] 
        
        \mshort & 6.1 & 1.2 & \textbf{75.6} & & 23.2 & 4.5 & \textbf{81.1} \\ 
        \bottomrule
    \end{tabular}
\end{table}

\begin{table}[t]
    \centering
    \small
    \renewcommand{\arraystretch}{1.1} 
    \setlength{\tabcolsep}{6pt}
    \caption{Experimental results on larger token size (\#token=784). We use ViT-T/8 as a backbone with an image size of $224^2$.  
    }
    \label{tab:longer_token_sequences}
    \vskip 0.15in
    \begin{tabular}{@{}lrrr@{}}
        \toprule
        
        Methods 
        & \specialcell{\#Param.\\(M)} 
        & \specialcell{FLOPs\\(G)}
        & \specialcell{Top-1\\Acc.(\%)} \\
        \midrule
        
        Softmax & 5.7 & 7.0 & 77.2 \\ 
        Performer & 5.7 & 4.9 & 67.2 \\ 
        Local attention & 5.7 & 4.4 & 70.6 \\
        Scatterbrain & 5.7 & 5.2 & 73.5 \\ 
        Nystromformer & 5.7 & 4.8 & 74.2 \\
        LARA & 5.8 & 4.6 & 75.0 \\ 
        Combiner & 5.7 & 4.7 & 75.6 \\ 
        Long-Short & 6.1 & 5.0 & 76.4 \\ 
        EVA & 5.8 & 4.6 & 76.7 \\ 
        \midrule
        
        \mshort~(ours) & 6.1 & 4.7 & \textbf{79.1} \\ 
        \bottomrule
    \end{tabular}
\end{table}

\begin{table}[t]
    \centering
    \small
    \renewcommand{\arraystretch}{1.1} 
    \setlength{\tabcolsep}{6pt}
    \caption{
    Experimental results on high-resolution input image sizes of [$448^2$, $672^2$, $896^2$].
    Each input image size yields token sizes of \#token=[784, 1764, 3136]. 
    OOM\textsuperscript{*} denotes out-of-memory with even an extremely small batch size of four.
    }
    \label{tab:beyond_the_limit_of_token_sequence}
    \vskip 0.15in
    \begin{tabular}{@{}llrrrrr@{}}
        \toprule

        Res
        & Methods 
        & \multicolumn{2}{c}{ViT-T/16} &
        & \multicolumn{2}{c}{ViT-S/16} \\ 
        \cmidrule(lr){3-4}
        \cmidrule(lr){6-7}

        &
        & \specialcell{FLOPs\\(G)}
        & \specialcell{Top-1\\Acc.(\%)}
        &
        & \specialcell{FLOPs\\(G)}
        & \specialcell{Top-1\\Acc.(\%)} \\ 
        \midrule

        \multirow{3}{*}{448\textsuperscript{2}}
        & Softmax & 7.1 & 76.7 && 22.5 & 83.1 \\ 
        & EVA & 4.8 & 75.9 && 18.0 & 82.1 \\ 
        & iMHSA & 4.8 & \textbf{78.6} && 17.9 & \textbf{83.1} \\ 
        \midrule

        \multirow{3}{*}{672\textsuperscript{2}}
        & Softmax & 24.5 & 77.2 && 67.6 & 83.5 \\ 
        & EVA & 10.9 & 76.1 && 40.5 & 82.2 \\ 
        & iMHSA & 10.7 & \textbf{79.1} && 40.3 & \textbf{83.5} \\ 
        \midrule

        \multirow{3}{*}{896\textsuperscript{2}}
        & Softmax & 62.4 & OOM\textsuperscript{*} && 158.1 & OOM\textsuperscript{*} \\ 
        & EVA & 19.2 & 76.1 && 71.8 & 82.5 \\ 
        & iMHSA & 19.0 & \textbf{79.4} && 71.6 & \textbf{83.7} \\ 
        \bottomrule
    \end{tabular}
\end{table}

\begin{table}[ht!]
    \scriptsize
    \definecolor{Gray}{gray}{0.9}
    \centering
    \renewcommand{\arraystretch}{1.2} 
    \setlength{\tabcolsep}{5pt}
    \caption{
    Experimental results of the image classification task on ImageNet-1k original Val~\cite{deng2009imagenet}, V2~\cite{recht2019imagenet}, and Real~\cite{beyer2020we} labels.
    We only include networks that are trained without extra data. 
    We report each network's speed on a single RTX 3090 GPU. 
    The Avg* denotes the average Top-1 accuracy of ImageNet's original val, V2, and real labels.
    We evaluate SOTA networks on V2 and Real labels using their public checkpoints.
    }
    \label{tab:comp-sota}
    \vskip 0.15in
    \begin{tabular}{@{}lrrrrrrr@{}}
    \toprule
    
    Network & 
    \multirow{2}{*}{\specialcell{Param.\\(M)}} & 
    \multirow{2}{*}{\specialcell{FLOPs\\(G)}} &
    \multirow{2}{*}{\specialcell{Speed\\(img/s)}} &
    \multicolumn{4}{c}{ImageNet Top1 Acc. (\%)} \\ 
    \cmidrule(lr){5-8}

    & & & & Val & V2 & Real & Avg\textsuperscript{*}\\
    \midrule

    Swin-T & 28.3 & 4.4 & 1676 & 81.3 & 69.5 & 86.7 & 79.1 \\ \hdashline[1pt/1pt]
    PoolFormer-S36 & 31.0 & 5.0 & 1156 & 81.4 & 69.9 & 86.6 & 79.3 \\ \hdashline[1pt/1pt]
    BiFormer-T & 13.1 & 2.2 & 1569 & 81.4 & 70.7 & 87.1 & 79.7 \\ \hdashline[1pt/1pt]
    CoAtNet-0 & 25.0 & 4.2 & 1635 & 81.6 & - & - & - \\ \hdashline[1pt/1pt] 
    ConvNeXt-T & 29.0 & 4.5 & 2040 & 82.1 & 71.3 & 87.3 & 80.2 \\ \hdashline[1pt/1pt] 
    Focal-T & 29.1 & 4.9 & 602 & 82.2 & - & - & - \\ \hdashline[1pt/1pt] 
    NAT-T & 28.0 & 4.3 & 1289 & 83.2 & 72.1 & 87.9 & 81.1 \\ \hdashline[1pt/1pt] 
    CoAtNet-1 & 42.2 & 8.4 & 918 & 83.3 & - & - & - \\ \hdashline[1pt/1pt] 
    InceptionNeXt-T & 28.1 & 4.2 & 1624 & 82.3 & 72.0 & 87.5 & 80.6 \\ \hdashline[1pt/1pt] 
    MogaNet-T & 25.0 & 5.0 & 805 & 83.4 & 72.7 & 88.0 & 81.4 \\ \hdashline[1pt/1pt] 
    MaxViT-T & 30.9 & 5.4 & 1009 & 83.6 & 72.9 & 88.0 & 81.5 \\ \hdashline[1pt/1pt] 
    iViT-T (Ours) & 27.5 & 4.0 & 1695 & \textbf{83.6} & \textbf{73.1} & \textbf{88.3} & \textbf{81.7} \\ 
    \midrule

    PoolFormer-M36 & 56.0 & 8.8 & 802 & 82.1 & 70.8 & 86.9 & 79.9 \\ \hdashline[1pt/1pt]
    Swin-S & 49.6 & 8.5 & 1017 & 83.0 & 71.8 & 87.7 & 80.8 \\ \hdashline[1pt/1pt] 
    ConvNeXt-S & 50.0 & 8.7 & 1257 & 83.1 & 72.4 & 88.1 & 81.2 \\ \hdashline[1pt/1pt] 
    Focal-S & 51.1 & 9.1 & 366 & 83.5 & - & - & - \\ \hdashline[1pt/1pt] 
    InceptionNeXt-S & 49.4 & 8.4 & 928 & 83.5 & 73.3 & 88.2 & 81.7 \\ \hdashline[1pt/1pt] 
    NAT-S & 51.0 & 7.8 & 834 & 83.7 & 73.2 & 88.1 & 81.7 \\ \hdashline[1pt/1pt] 
    BiFormer-S & 26.0 & 4.5 & 745 & 83.8 & 73.6 & 88.3 & 81.9 \\ \hdashline[1pt/1pt] 
    CoAtNet-2 & 74.7 & 15.7 & 613 & 84.1 & - & - & - \\ \hdashline[1pt/1pt] 
    MogaNet-S & 44.0 & 9.9 & 400 & 84.3 & 74.0 & 88.5 & 82.3 \\ \hdashline[1pt/1pt]
    MaxViT-S & 69.0 & 11.7 & 654 & 84.5 & 73.9 & 88.5 & 82.3 \\ \hdashline[1pt/1pt] 
    iViT-S (Ours) & 41.1 & 7.8 & 930 & \textbf{84.5} & \textbf{74.6} & \textbf{88.8} & \textbf{82.7} \\ 
    \midrule

    PoolFormer-M48 & 73.0 & 11.6 & 603 & 82.5 & 71.6 & 87.2 & 80.4 \\ \hdashline[1pt/1pt] 
    Swin-B & 87.7 & 15.1 & 726 & 83.5 & 72.3 & 87.9 & 81.2 \\ \hdashline[1pt/1pt] 
    Focal-B & 89.8 & 16.0 & 273 & 83.8 & - & - & - \\ \hdashline[1pt/1pt]
    ConvNeXt-B & 89.0 & 15.4 & 886 & 83.8 & 73.7 & 88.3 & 81.9 \\ \hdashline[1pt/1pt] 
    InceptionNeXt-B & 86.7 & 14.8 & 640 & 84.0 & 74.1 & 88.4 & 82.1 \\ \hdashline[1pt/1pt] 
    BiFormer-B & 57.0 & 9.8 & 483 & 84.3 & 74.0 & 88.5 & 82.3 \\ \hdashline[1pt/1pt]
    NAT-B & 90.0 & 13.7 & 614 & 84.3 & 74.1 & 88.6 & 82.3 \\ \hdashline[1pt/1pt] 
    CoAtNet-3 & 167.0 & 34.7 & 347 & 84.5 & - & - & - \\ \hdashline[1pt/1pt] 
    NFNet-F1 & 132.6 & 35.5 & 239 & 84.7 & 71.9 & 88.1 & 81.6 \\ \hdashline[1pt/1pt] 
    MogaNet-B & 83.0 & 15.9 & 284 & 84.7 & 74.1 & 88.4 & 82.4 \\ \hdashline[1pt/1pt] 
    MaxViT-B & 120.0 & 23.4 & 361 & 85.0 & 74.3 & 88.6 & 82.6 \\ \hdashline[1pt/1pt] 
    iViT-B (Ours) & 58.3 & 13.1 & 686 & \textbf{85.1} & \textbf{75.2} & \textbf{89.0} & \textbf{83.1} \\ 
    \bottomrule
  \end{tabular}
\end{table}

For the results presented in \Cref{tab:baseline_networks,tab:longer_token_sequences,tab:beyond_the_limit_of_token_sequence}, 
all methods are trained with the same hyperparameters.
\Cref{tab:baseline_networks} shows the experimental results of Vision Transformers with an input image size of $224^2$ and a patch size of $16$, resulting in a token size of $(224/16)^2=196$.
The proposed iMHSA performs favorably against other efficient attention methods using manageable resources.
To showcase the effectiveness under the larger token size, we use a smaller path size of $8$ in \Cref{tab:longer_token_sequences} and a larger input size in \Cref{tab:beyond_the_limit_of_token_sequence}. 
Experimental results of \Cref{tab:longer_token_sequences,tab:beyond_the_limit_of_token_sequence} show that the previous efficient attention methods exhibit considerably reduced computational complexity compared to the baseline softmax, but almost the same or worse performance of Top-1 classification accuracy.
On the other hand, the proposed iMHSA maintains the trend of lower computational complexity but with much higher classification performance.

\subsection{Evaluation against State-of-the-art Transformers}

We compare the proposed iViT with other state-of-the-art models, such as 
Swin~\cite{liu2021swin}, 
Focal~\cite{yang2021focal},
CoAtNet~\cite{dai2021coatnet}, 
ConvNeXt~\cite{liu2022convnet}, 
MaxViT~\cite{tu2022maxvit}, 
PoolFormer~\cite{yu2022metaformer},
InceptionNeXt~\cite{yu2023inceptionnext}, 
BiFormer~\cite{zhu2023biformer},
NFNet~\cite{brock2021high},
MogaNet~\cite{li2024efficient},
and NAT~\cite{hassani2023neighborhood}.

\subsubsection{Image Classification}
We evaluate the proposed iViT on the ImageNet-1K dataset. 
We train our models by 300 epochs employing standard data augmentation methods such as CutMiX~\cite{yun2019cutmix}, MixUp~\cite{zhang2017mixup}, and Rand-Augmentation~\cite{cubuk2020randaugment}.\footnote{The detailed hyperparameter configurations can be found in the 
Appendix.}
\Cref{tab:comp-sota} shows that our iViT performs favorably in terms of accuracy and speed trade-off across all model scales on the ImageNet-1k and its out-of-distribution datasets.

\subsubsection{Object Detection and Instance Segmentation}
We compare our iViT with SOTA networks on the COCO \cite{lin2014microsoft}.
We use Mask-RCNN as a backbone and train it with $\times 3$ schedule. 
As a result, the proposed iViT demonstrates competitive performance using an even smaller model size as shown in \Cref{tab:object_instance}.

\subsubsection{Semantic Segmentation}
We perform experimental comparison for a semantic segmentation task on the ADE20K~\cite{zhou2017scene}.
We utilize UperNet~\cite{xiao2018unified} as a backbone with a $160K$ training schedule. \Cref{tab:semantic} presents that the proposed iViT considerably improves the segmentation performance using minimal parameters.

\begin{table}[t]
    \centering
    \scriptsize
    \renewcommand{\arraystretch}{1.2} 
    \setlength{\tabcolsep}{4pt}
    \caption{
        Experimental results of the object detection and instance segmentation tasks on COCO2017~\cite{lin2014microsoft} dataset. We use Mask-RCNN~\cite{cai2018cascade} as the detection network.
        Following Swin~\cite{liu2021swin} and ConvNext~\cite{liu2022convnet}, we measure FLOPs with the input image size of $1280 \times 800$.
    }
    \label{tab:object_instance}
    \vskip 0.15in
    \begin{tabular}{@{}
    lrr
    r>{\scriptsize}r>{\scriptsize}r
    r>{\scriptsize}r>{\scriptsize}r
    @{}}
        \toprule

        Network
        & \specialcell{Param.\\(M)} 
        & \specialcell{FLOPs\\(G)} 
        & \specialcell{$\text{AP}^{\text{b}}$\\(\%)}
        & \specialcell{$\text{AP}_{\text{50}}^{\text{b}}$\\(\%)}
        & \specialcell{$\text{AP}_{\text{75}}^{\text{b}}$\\(\%)}
        & \specialcell{$\text{AP}^{\text{m}}$\\(\%)}
        & \specialcell{$\text{AP}_{\text{50}}^{\text{m}}$\\(\%)} 
        & \specialcell{$\text{AP}_{\text{75}}^{\text{m}}$\\(\%)} 
        \\
        \midrule

        Swin-T & 48 & 267 & 46.0 & 68.1 & 50.3 & 41.6 & 65.1 & 44.9 \\ \hdashline[1pt/1pt]  
        ConvNeXt-T  & 48 & 262 & 46.2 & 67.9 & 50.8 & 41.7 & 65.0 & 44.9 \\ \hdashline[1pt/1pt]  
        PVTv2-B2  & 45 & 309 & 47.8 & 69.7 & 52.6 & 43.1 & 66.8 & 46.7 \\ \hdashline[1pt/1pt]  
        Focal-T & 49 & 291 & 47.2 & 69.4 & 51.9 & 42.7 & 66.5 & 45.9 \\ \hdashline[1pt/1pt] 
        ViT-Adapter-S & 48 & 403 & 48.2 & 69.7 & 52.5 & 42.8 & 66.4 & 45.9 \\ \hdashline[1pt/1pt]  
        iViT-T (ours) & 46 & 257 & \textbf{48.7} & \textbf{70.9} & \textbf{53.4} & \textbf{44.1} & \textbf{68.0} & \textbf{47.4} \\ 
        \midrule
        
        Swin-S & 69 & 354 & 48.2 & 69.8 & 52.8 & 43.2 & 67.0 & 46.1 \\ \hdashline[1pt/1pt]  
        ConvNeXt-S  & 70 & 348 & 47.9 & 70.0 & 52.7 & 42.9 & 66.9 & 46.2 \\ \hdashline[1pt/1pt] 
        PVTv2-B3  & 65 & 397 & 48.4 & 69.8 & 53.3 & 43.2 & 66.9 & 46.7 \\ \hdashline[1pt/1pt]  
        Focal-S & 71 & 401 & 48.8 & 70.5 & 53.6 & 43.8 & 67.7 & 47.2 \\ \hdashline[1pt/1pt] 
        iViT-S (ours) & 60 & 332 & \textbf{49.8} & \textbf{71.8} & \textbf{54.6} & \textbf{44.7} & \textbf{69.0} & \textbf{48.3} \\ 
        \midrule
        
        Swin-B & 107 & 496 & 48.6 & 70.0 & 53.4 & 43.3 & 67.1 & 46.7 \\ \hdashline[1pt/1pt]  
        ConvNeXt-B  & 108 & 486 & 48.5 & 70.1 & 53.3 & 43.5 & 67.1 & 46.7 \\ \hdashline[1pt/1pt]  
        PVTv2-B5  & 102 & 557 & 48.4 & 69.2 & 52.9 & 42.9 & 66.6 & 46.2 \\ \hdashline[1pt/1pt]  
        Focal-B & 110 & 533 & 49.0 & 70.1 & 53.6 & 43.7 & 67.6 & 47.0 \\ \hdashline[1pt/1pt] 
        ViT-Adapter-B  & 120 & 832 & 49.6 & 70.6 & 54.0 & 43.6 & 67.7 & 46.9 \\ \hdashline[1pt/1pt]  
        iViT-B (ours) & 77 & 442 & \textbf{50.6} & \textbf{72.3} & \textbf{55.8} & \textbf{45.1} & \textbf{69.6} & \textbf{49.0} \\ 
        \bottomrule
    \end{tabular}
\end{table}

\begin{table}[t]
    \centering
    \small
    \renewcommand{\arraystretch}{1.2} 
    \setlength{\tabcolsep}{6pt}
    \caption{
        Experimental results of the semantic segmentation task on ADE20K~\cite{zhou2017scene}. We use UperNet~\cite{xiao2018unified} as the segmentation head network. We report the results of the 160K schedule. Following ConvNext~\cite{liu2022convnet}, we report the multi-crop mIOU results and measure FLOPs with the input image size of $2048 \times 512$.
    }
    \label{tab:semantic}
    \vskip 0.15in
    \begin{tabular}{@{}lrrr@{}}
        \toprule
        
        Network
        & \specialcell{Param.\\(M)} 
        & \specialcell{FLOPs\\(G)} 
        & \specialcell{mIOU\\(\%)} \\
        \midrule

        Swin-T & 60 & 941 & 45.8 \\ 
        ConvNeXt-T  & 60 & 939 & 46.7 \\ 
        Focal-T & 62 & 998 & 47.0 \\ 
        iViT-T (ours) & 56 & 940 & \textbf{49.7} \\ 
        \midrule
        
        Swin-S & 81 & 1032 & 49.5 \\ 
        ConvNeXt-S  & 82 & 1027 & 49.6 \\ 
        Focal-S & 85 & 1027 & 50.0 \\ 
        iViT-S (ours) & 74 & 1029 & \textbf{51.4} \\ 
        \midrule
        
        Swin-B & 121 & 1188 & 49.7 \\ 
        ConvNeXt-B  & 122 & 1170 & 49.9 \\ 
        Focal-B & 126 & 1354 & 50.5 \\ 
        iViT-B (ours) & 88 & 1142 & \textbf{52.5} \\ 
        \bottomrule
    \end{tabular}
\end{table}

\begin{table}[t]
    \small
    \centering
    \renewcommand{\arraystretch}{1.1} 
    \setlength{\tabcolsep}{6pt}
    \caption{Experimental results of the factor analysis using ViT-Tiny on the ImageNet dataset.
    Any of the two proposed methods significantly reduces the performance in terms of runtime or accuracy.
    }
    \label{tab:comp_wise_effect}
    \vskip 0.15in
    \begin{tabular}{
        ccrrr
    }
        \toprule
        \specialcell{\scriptsize Self-Attention \\ \scriptsize decomposition} & 
        \specialcell{\scriptsize Cross-head \\ \scriptsize interaction} &
        \specialcell{FLOPs\\(G)} & 
        \specialcell{Runtime\\(ms)} & 
        \specialcell{Top-1\\Acc.(\%)} \\ 
        \midrule
        \checkmark & \checkmark & 3.3 & 23 & \textbf{74.2} \\ 
        \checkmark & \xmark & 3.3 & \textbf{19} & 71.4 \\ 
        \xmark & \checkmark & 5.8 & 76 & 74.2\\ 
        \bottomrule
    \end{tabular}
\end{table}

\subsection{Ablation Study}
\label{sec:ablation}

We evaluate the efficacy of each element of the proposed method.
We conduct an element-wise evaluation on the ImageNet-1K dataset. As shown in \cref{tab:comp_wise_effect}, each element of the proposed method contributes to improving the performance of its respective baseline.

\subsection{Analysis}
\label{sec:analysis}

For our analytic studies, we use the CIFAR100~\cite{krizhevsky2009learning} dataset with ViT-Tiny/Small network.

\begin{figure}[tbh!]
    \centering
    \begin{subfigure}[b]{0.36\paperwidth}
        \centering        
        \includegraphics[width=\linewidth]{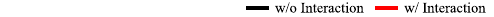}
    \end{subfigure}

    \begin{subfigure}[b]{0.36\paperwidth}
        \centering        
        \includegraphics[width=0.49\linewidth]{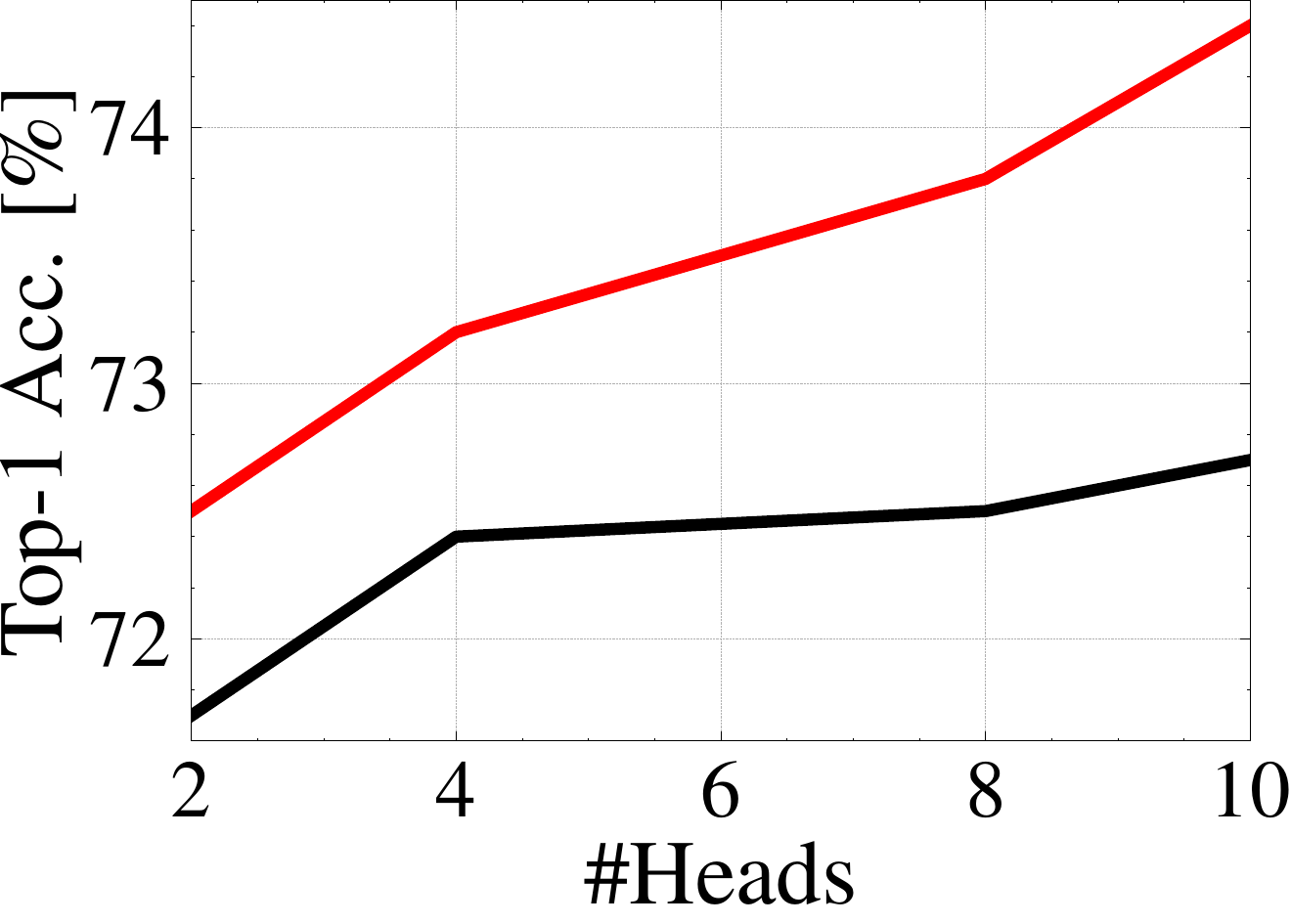}
        \centering        
        \includegraphics[width=0.49\linewidth]{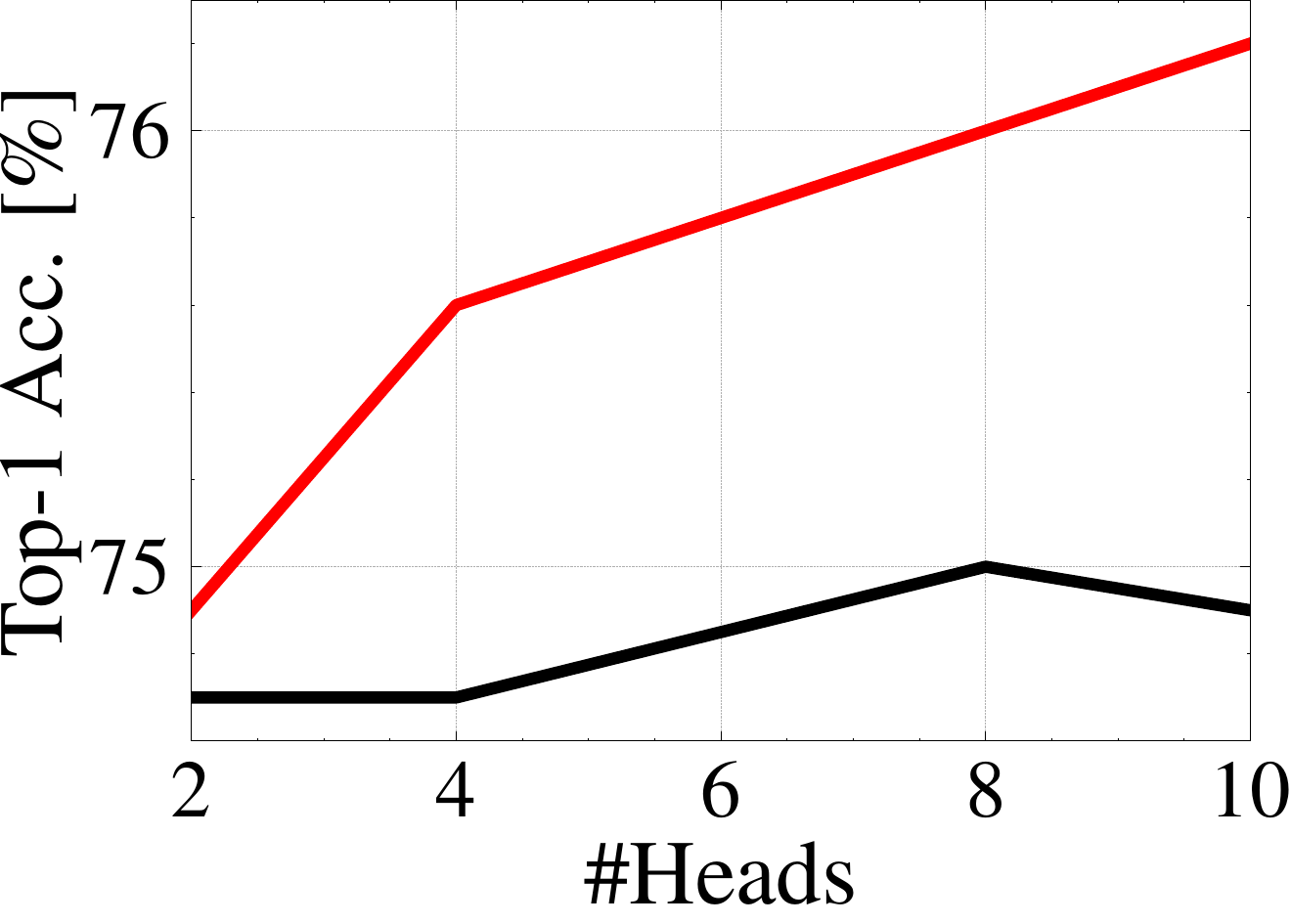}
        \caption{Accuracy}
        \label{fig:plot_head_acc}
    \end{subfigure}

    \begin{subfigure}[b]{0.36\paperwidth}
        \centering        
        \includegraphics[width=0.49\linewidth]{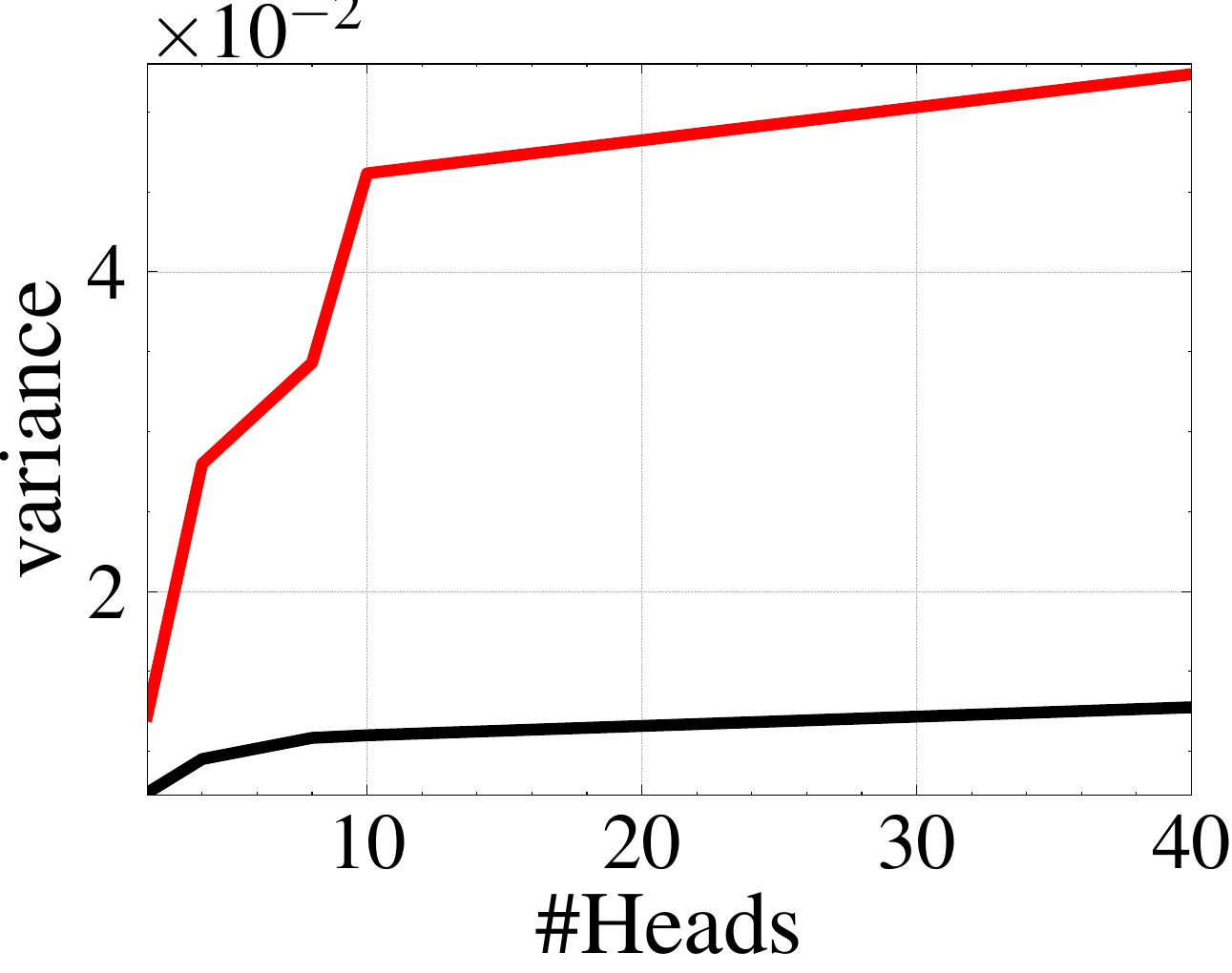}
        \includegraphics[width=0.49\linewidth]{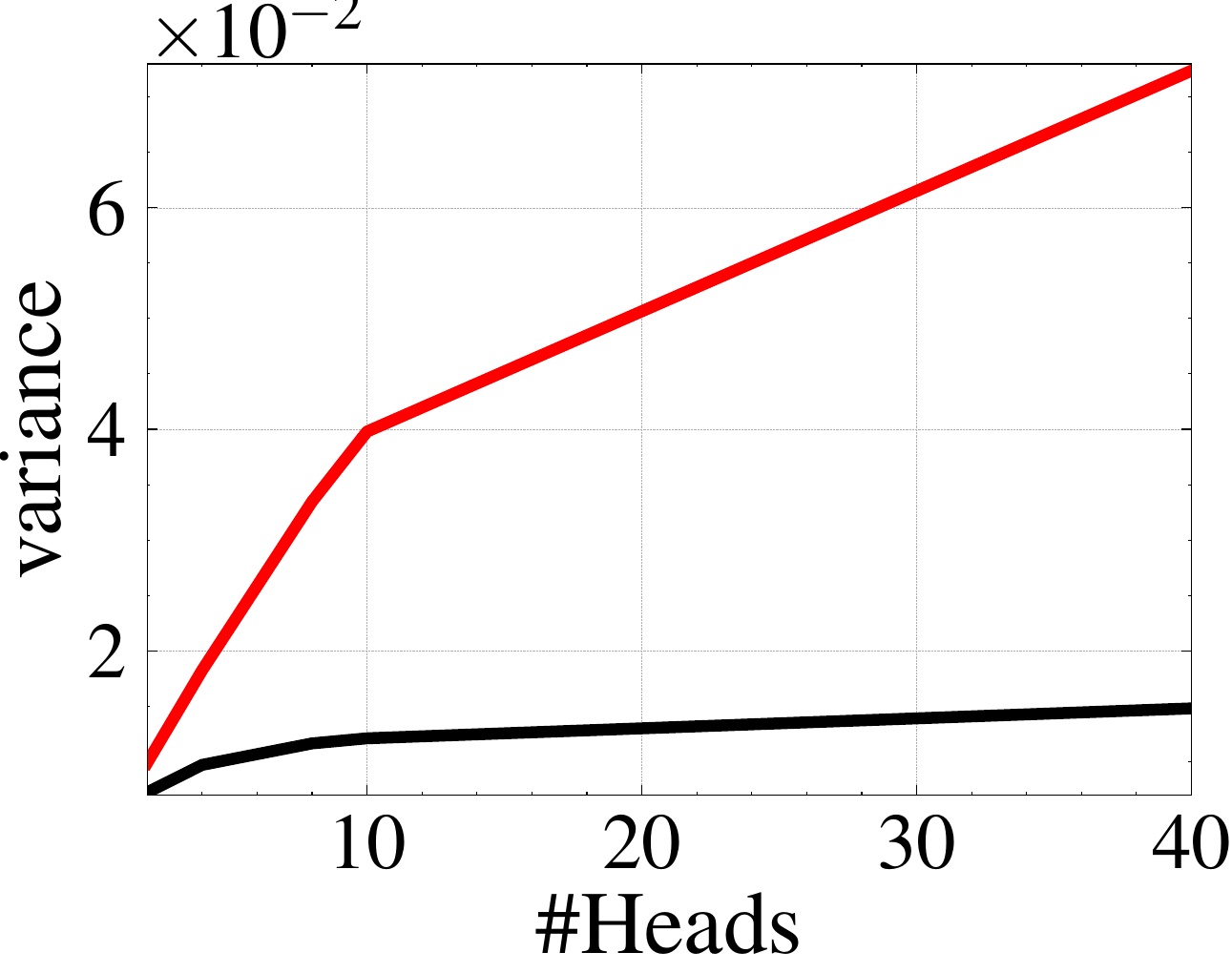}
        \caption{Variance}
        \label{fig:plot_head_variance}
    \end{subfigure}
    
    \caption{
    Experimental results of top-1 accuracy and the variance of attention matrices depending on head numbers. We use Tiny(left) and Small(right)-scale architectures of ViTs.  
    We estimate the variance of the attention matrices in each head dimension across different layers in a network.
    }
    \label{fig:plot_head_acc_variance}
\end{figure}

\paragraph{Number of Heads.} We evaluate the performance of the proposed interaction method with regard to the number of heads in the ViT model. \cref{fig:plot_head_acc} shows that, in the absence of the interaction, performance reaches a plateau rapidly. However, with our interaction method, performance is improved consistently as the number of heads increases. This finding is significant as it can suggest a new direction for designing the multi-head self-attention layers. Although we do not prioritize determining the best head numbers in designing our iViT, we believe that it will serve as an important ground for future studies in developing the architecture of the MHSA.

\begin{figure}[tbh!]
    \centering
    \begin{subfigure}[b]{0.36\paperwidth}
        \centering        
        \includegraphics[width=\linewidth]{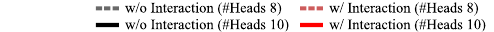}
    \end{subfigure}

    \begin{subfigure}[b]{0.18\paperwidth}
        \centering        
        \includegraphics[width=\linewidth]{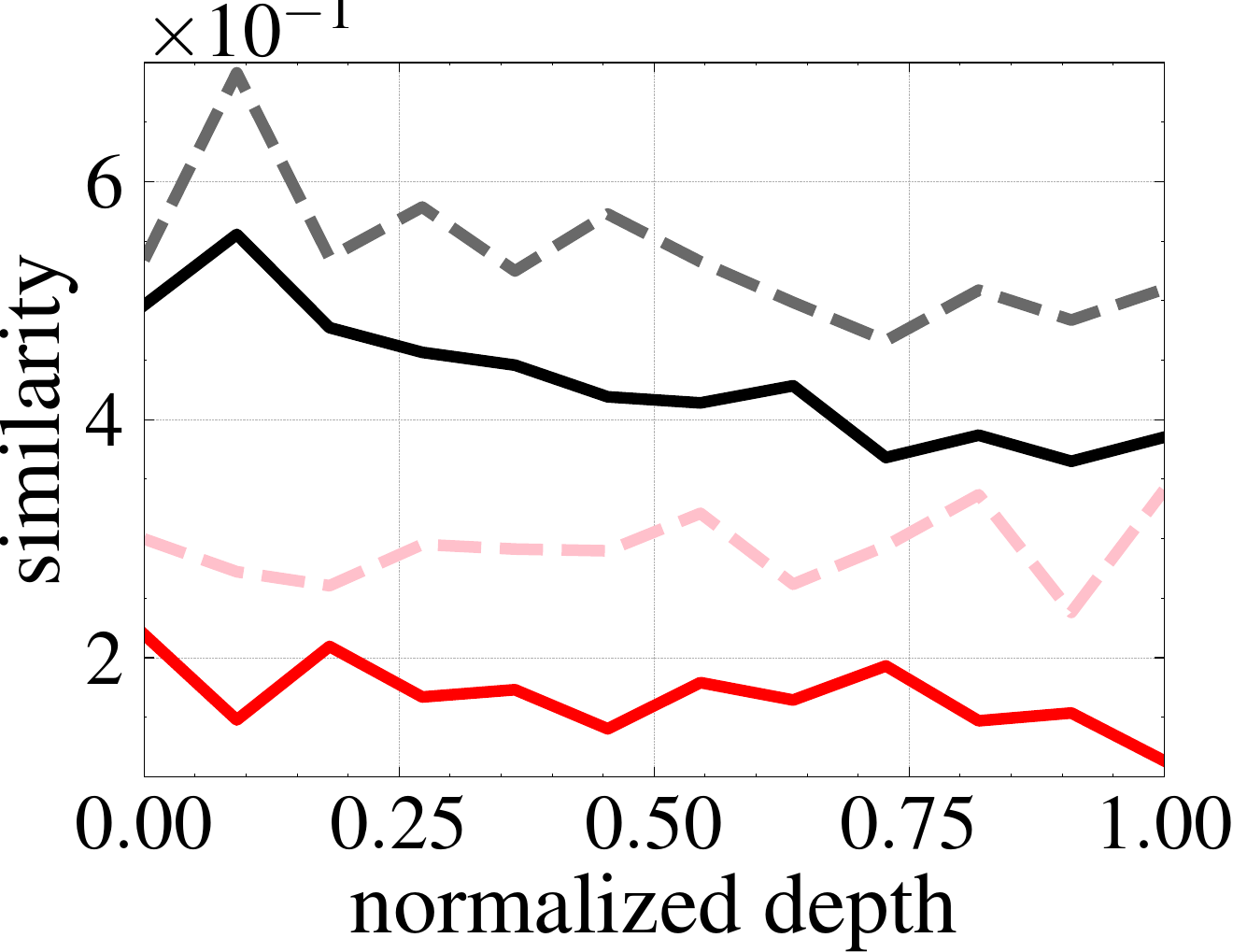}
        \caption{Tiny}
        \label{fig:plot_depth_attention_similarity_tiny}
    \end{subfigure}
    \begin{subfigure}[b]{0.18\paperwidth}
        \centering        
        \includegraphics[width=\linewidth]{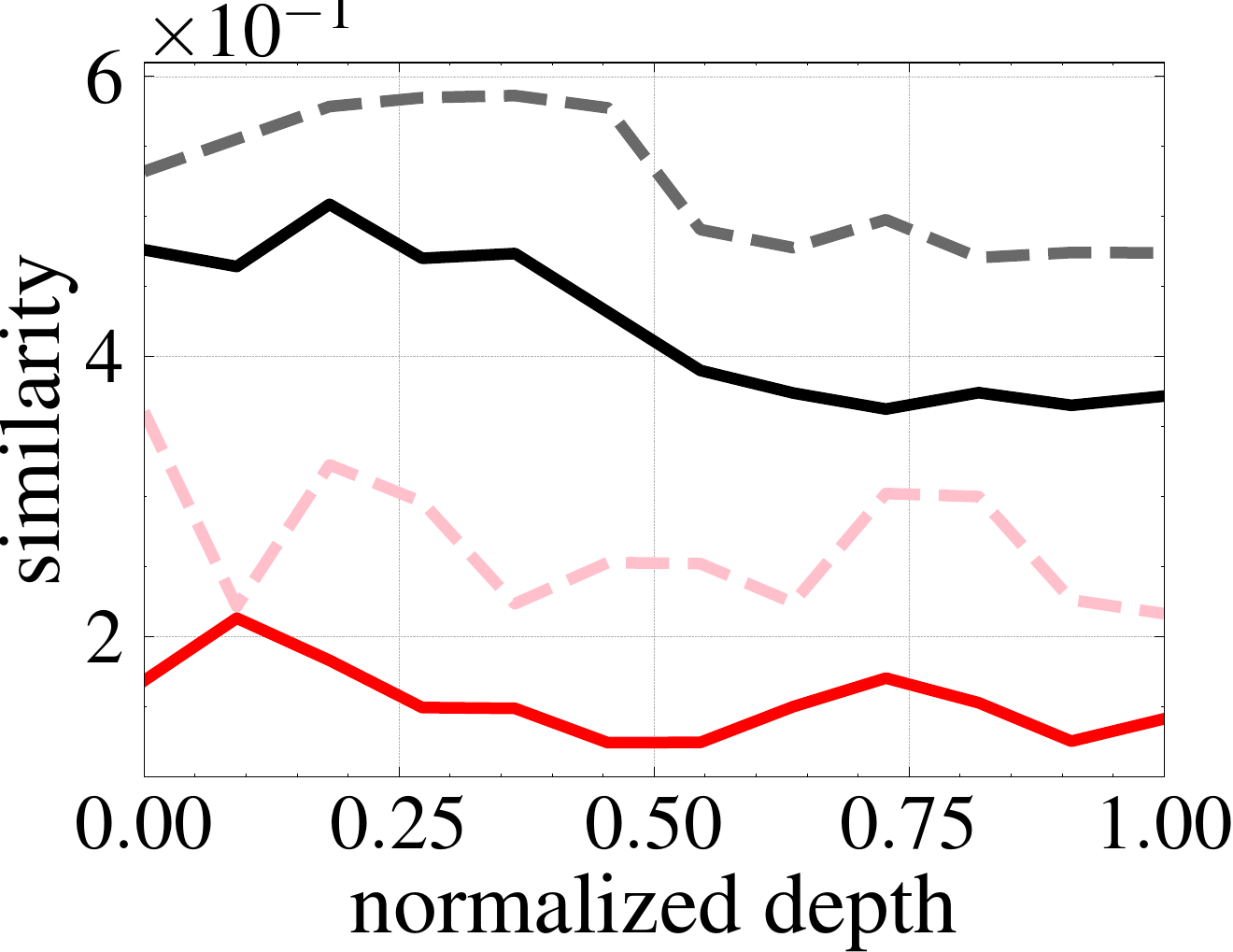}
        \caption{Small}
        \label{fig:plot_depth_attention_similarity_small}
    \end{subfigure}
    
    \caption{
    Experimental results of the similarity of attention matrices across different depths in a network.
    We measure the cosine similarity between attention matrices in different multi-heads.
    }
    \label{fig:plot_depth_attention_similarity}
\end{figure}

\paragraph{Feature Diversity.} 
At the beginning of this paper, we assume that establishing strong interactions in networks provides the benefit of learning diversified feature distribution.
To confirm this assumption, we validate the improvement of the feature diversity in networks by the proposed interaction approach.
In this validation, we leverage the previous works~\cite{park2021vision,zhou2021refiner}, which indicates that the high variance and low similarity of a network improve the feature diversity.
Thus, we measure the variance and similarity as the indicator of the feature diversity as shown in \cref{fig:plot_depth_attention_similarity,fig:plot_head_variance}.
They confirm our assumption as the proposed interaction approach shows consistent trends of high variance and low similarity.

\begin{figure*}[tbh!]
    \centering
    \begin{subfigure}[b]{0.67\paperwidth}
        \centering        
        \includegraphics[width=\linewidth]{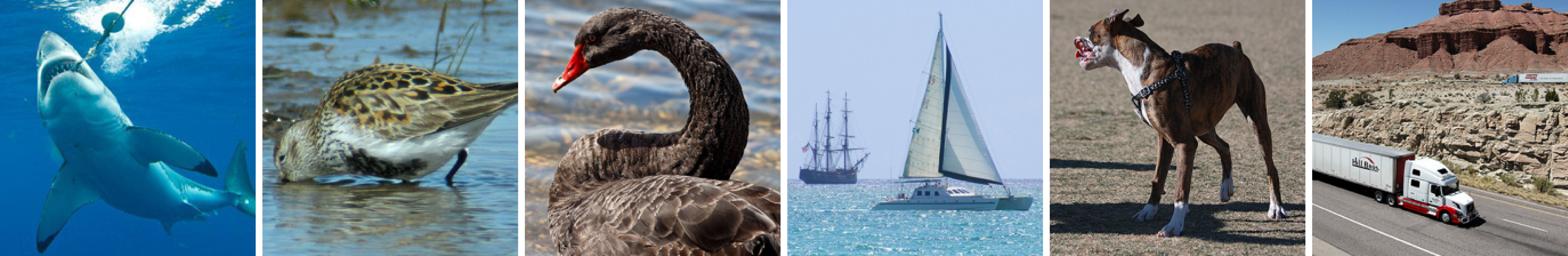}
        \subcaption{Image}
        \label{fig:fig_attn_img}
    \end{subfigure}

    \begin{subfigure}[b]{0.67\paperwidth}
        \centering        
        \includegraphics[width=\linewidth]{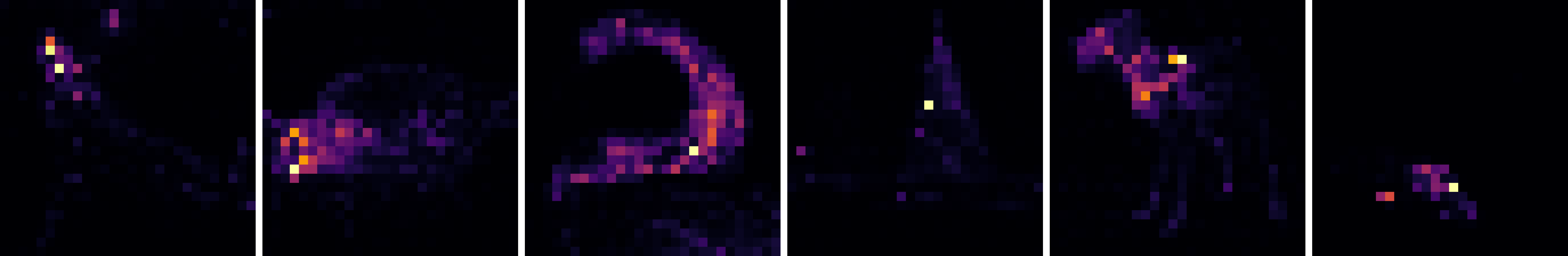}
        \subcaption{MHSA}
        \label{fig:fig_attn_mhsa}
    \end{subfigure}

    \begin{subfigure}[b]{0.67\paperwidth}
        \centering        
        \includegraphics[width=\linewidth]{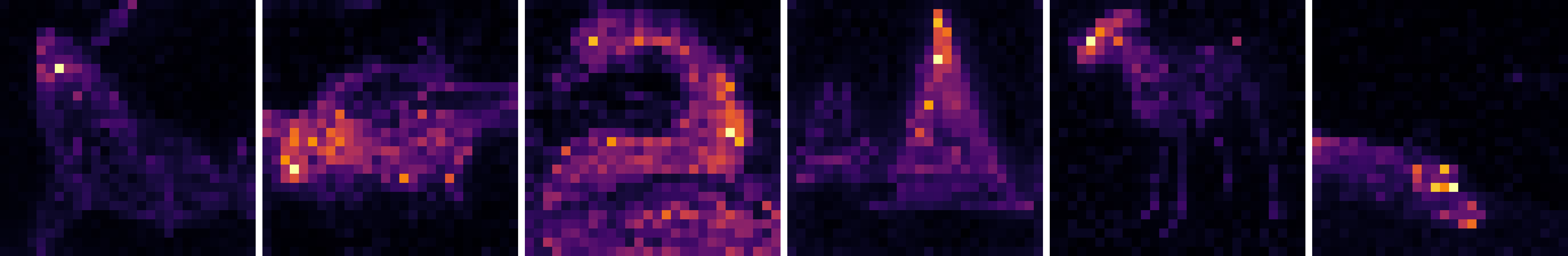}
        \subcaption{iMHSA}
        \label{fig:fig_attn_imhsa}
    \end{subfigure}
    
    \caption{
    Visualization of attention matrix of the original MHSA and our iMHSA method. We use the ViT-Tiny/8 model trained on the ImageNet-1K dataset. 
    }
    \label{fig:attn_map_visualization}
\end{figure*}

\subsection{Qualitative Results}
We provide qualitative results of our method as shown in \cref{fig:attn_map_visualization}. The proposed iMHSA effectively captures the object boundary compared to the original MHSA. This finding provides further evidence that our interaction approach enhances the diversity of feature representation in an attention operation.  

\section{Conclusion}
This work aims to facilitate the information flow between multi-head attention matrices with linear complexity. 
To achieve this, we introduce the iMHSA, which enables interactions across different heads. 
Due to the quadratic computational cost of the interaction layer, we reduce the dimension of the attention matrix by decomposing it into query- and key-less components. 
Through comprehensive experiments, we demonstrate that the proposed method achieves competitive performance while minimizing computing costs on various visual recognition tasks.
We show that our method is effective with large token sizes of high-resolution images compared to the previous efficient attention methods.
We evaluate the benefits of our interaction approach in terms of head numbers, variance, and similarity in a network.  
We anticipate that our method can serve as a foundation for future study.

\newpage

\section{Broader Impact}
We introduce a new architectural method for better performance and computational complexity of the machine learning model.
We believe our paper does not raise any significant social impact concerns.

\bibliography{paper}
\bibliographystyle{icml2024}

\appendix
\onecolumn

\section{Hyper-Parameter Specification}
We provide details of the hyper-parameter settings on the ImageNet-1K~\cite{deng2009imagenet}, MSCOCO-2017~\cite{lin2014microsoft}, and ADE20K~\cite{zhou2017scene} datasets. 

\subsection{Image Classification}
We use two backbone networks for the image classification task.
In the ablation study, we utilize the plain ViT model for quick training while employing our new backbone, iVIT, in comparison on the SOTA networks.

\paragraph{Plain ViT.} For a fair comparison with previous studies, we follow the ImageNet-1K hyper-parameter of ~\citet{touvron2021training} as shown in \cref{tab:imagenet_1k_hyperparameter} for plain ViT architecture in~\cref{tab:baseline_networks,tab:longer_token_sequences}. We use 30 epoch setting as shown in \cref{tab:imagenet_1k_hyperparameter} for fine-tuning our networks on high-resolution images.

\paragraph{Our iViT.} For training our iViT in~\cref{tab:comp-sota}, we use the similar hyper-parameters of ~\citet{yu2022metaformer} as shown in \cref{tab:imagenet_1k_hyperparameter}. Specifically, we adopt commonly used augmentation methods such as CutMix~\cite{yun2019cutmix}, MixUp~\cite{zhang2017mixup}, and RandAugmentation~\cite{cubuk2020randaugment}. 

\subsection{Object Detection and Instance Segmentation}
We utilize the Mask R-CNN~\cite{he2017mask} head provided in MMDetection~\cite{mmdetection} for object detection and instance segmentation tasks. We adopt the hyper-parameter setting of Swin~\cite{liu2021swin}. In detail, as shown in~\cref{tab:coco_ade_hyperparameter}, we train our network for 36 epochs with an initial learning rate of 1e-4, lr-scheduler of step-lr-decay, augmentation of random-resized-crop and horizontal flip, and optimizer of AdamW.

\subsection{Semantic Segmentation}
We exploit the UperNet head~\cite{xiao2018unified} provided in MMSegmentation~\cite{mmseg2020} for the semantic segmentation task. We follow the hyper-parameter of Swin~\cite{liu2021swin}. Especially, as shown in~\cref{tab:coco_ade_hyperparameter}, we train our network for 160K iteration with an initial learning rate of 5e-8, lr-scheduler of polynomial-lr-decay, and optimizer of AdamW. Moreover, we adopt multi-scale augmentation during inference time.

\begin{table}[bh]
    \centering
        \begin{tabular}{lccc}
            \toprule
            & ~\cref{tab:baseline_networks,tab:longer_token_sequences} 
            & ~\cref{tab:beyond_the_limit_of_token_sequence}
            & ~\cref{tab:comp-sota} \\ 
            \midrule
    
            network &
            ViT-T/S &
            ViT-T/S &
            iViT-T/S/B \\
            image size 
            & 224\textsuperscript{2} 
            & 448\textsuperscript{2} / 672\textsuperscript{2} / 896\textsuperscript{2} 
            & 224 \\ 
            epochs & 300 & 30 & 300 \\
            \midrule
            
            batch size & 1024 & 1024 & 4096 \\ 
            optimizer & AdamW & AdamW & LAMB \\ 
            learning rate & 1e-3 & 5e-6 & 8e-3 \\ 
            learning rate decay & cosine & \xmark & cosine \\ 
            weight decay & 5e-2 & 1e-8 & 5e-2 \\ 
            warmup epochs & 5 & 5 & 20 \\ 
            label smoothing & 0.1 & 0.1 & 0.1 \\ 
            stoch. depth & 0.1 & 0.1 & 0.15/0.3/0.4 \\ 
            \midrule
            
            rand augment & \multicolumn{3}{c}{9 / 0.5} \\         
            mixup prob. & \multicolumn{3}{c}{0.8} \\ 
            cutmix prob. & \multicolumn{3}{c}{1} \\ 
            erasing prob. & \multicolumn{3}{c}{0.25} \\ 
            \bottomrule
    \end{tabular}
    \caption{ImageNet hyperparameter specifications for each table in the main paper.}
    \label{tab:imagenet_1k_hyperparameter}
\end{table}

\clearpage

\begin{table}[!th]
    \centering
    \begin{subtable}[t]{0.3\textwidth}
        \begin{tabular}{lc}
            \toprule
             & \cref{tab:object_instance} \\
            \midrule
            
            network & iViT-T/S/B \\
            image size & (800, 1333) \\
            epoch & 36 \\
            \midrule
            batch size & 16 \\ 
            optimizer & AdamW \\ 
            learning rate & 1e-4 \\
            learning rate decay & step-lr \\
            weight decay & 5e-2 \\
            stoch. depth & 0.2/0.3/0.4 \\
            \bottomrule
        \end{tabular}
        \subcaption{COCO-2017}
        \label{tab:coco_hyperparameter}
    \end{subtable}
    \hspace{4pt}
    \begin{subtable}[t]{0.3\textwidth}
        \begin{tabular}{lc}
            \toprule
             & \cref{tab:semantic} \\
            \midrule
            
            network & iViT-T/S/B \\
            image size & (512, 2048) \\
            iteration & 160K \\
            \midrule
            batch size & 16 \\ 
            optimizer & AdamW \\ 
            learning rate & 8e-5 \\
            learning rate decay & poly \\
            weight decay & 5e-2 \\
            stoch. depth & 0.4 \\
            \bottomrule
        \end{tabular}
        \subcaption{ADE20K}
        \label{tab:ade_hyperparameter}
    \end{subtable}
    \caption{COCO-2017 and ADE20K hyperparameter specifications for each table in the main paper.}
    \label{tab:coco_ade_hyperparameter}
\end{table}

\end{document}